\documentclass{article}

\usepackage{arxiv}

\usepackage[utf8]{inputenc} 
\usepackage[T1]{fontenc}    
\usepackage[colorlinks=true]{hyperref}       
\hypersetup{
	colorlinks,
	citecolor=black,
	linkcolor=black,
	urlcolor=black}
\usepackage[caption=false]{subfig}
\usepackage{url}            
\usepackage{booktabs}       
\usepackage{amsfonts}       
\usepackage{nicefrac}       
\usepackage{microtype}      
\usepackage{amsmath}
\usepackage{cleveref}       
\usepackage{graphicx}
\usepackage[numbers,sort&compress]{natbib}
\usepackage{doi}

\usepackage{multirow}
\usepackage{multicol}

\usepackage{adjustbox}
\newsavebox\CBox
\def\textBF#1{\sbox\CBox{#1}\resizebox{\wd\CBox}{\ht\CBox}{\textbf{#1}}}

\usepackage{xcolor}
\usepackage[ruled,linesnumbered]{algorithm2e}
\newlength\mylen

\SetKwComment{Comment}{$\triangleright$\ }{}
\SetKwInOut{KwIn}{Input}
\SetKwInOut{KwOut}{Output}

\SetCommentSty{mycommfont}
\title{A Deep Convolutional Neural Network for Salt-and-pepper Noise Removal Using Selective Convolutional Blocks}

\date{}

\author{ \href{https://orcid.org/0000-0002-6413-7925}{\includegraphics[scale=0.07]{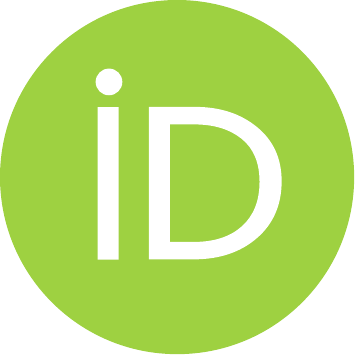}}\hspace{1mm}Ahmad Ali Rafiee\\
	Department of Electronics and \\Telecommunications Engineering\\
	Shiraz University\\
	Shiraz, Iran\\
	\href{mailto:aa.rafiee@shirazu.ac.ir}{\texttt{aa.rafiee@shirazu.ac.ir}} \\
	\And
	\href{https://orcid.org/0000-0001-9383-8475}{\includegraphics[scale=0.07]{orcid.pdf}}\hspace{1mm}Mahmoud Farhang\\
	Department of Electronics and \\Telecommunications Engineering\\
	Shiraz University\\
	Shiraz, Iran\\
	\href{mailto:mfarhang@shirazu.ac.ir}{\texttt{mfarhang@shirazu.ac.ir}} \\
}

\renewcommand{\headeright}{}
\renewcommand{\shorttitle}{A Deep Convolutional Neural Network for Salt-and-pepper Noise Removal Using Selective Convolutional Blocks}

\hypersetup{
pdftitle={A Deep Convolutional Neural Network for Salt-and-pepper Noise Removal Using Selective Convolutional Blocks},
pdfauthor={Ahmad Ali Rafiee, Mahmoud Farhang},
}

\begin{document}
\maketitle

\begin{abstract}
	In recent years, there has been an unprecedented upsurge in applying deep learning approaches, specifically convolutional neural networks (CNNs), to solve image denoising problems, owing to their superior performance. However, CNNs mostly rely on Gaussian noise, and there is a conspicuous lack of exploiting CNNs for salt-and-pepper (SAP) noise reduction. In this paper, we proposed a deep CNN model, namely SeConvNet, to suppress SAP noise in gray-scale and color images. To meet this objective, we introduce a new selective convolutional (SeConv) block.
	SeConvNet is compared to state-of-the-art SAP denoising methods using extensive experiments on various common datasets. The results illustrate that the proposed SeConvNet model effectively restores images corrupted by SAP noise and surpasses all its counterparts at both quantitative criteria and visual effects, especially at high and very high noise densities.
\end{abstract}


\section{Introduction}
\label{Introduction}

Images are polluted by miscellaneous types of noises during acquisition, compression, and transmission. The resulting image information loss and poor quality of images give rise to considerable difficulties for many other image processing and computer vision tasks  \cite{Bovik2005, Bhadouria2014, Ghimpeteanu2016}. Therefore, image denoising is considered a classical yet still vital low-level vision issue, and is a basic necessity in many computer vision applications.  The image denoising process aims to remove noise from noisy images along with retaining the edges, textures, and details of images, which leads to obtaining denoised images with superior quality \cite{Xu2017}. Impulse noise is one of the prevalent types of noise in images that changes the value of some randomly selected pixels in images. Impulse noise can be classified into two categories, i.e., salt-and-pepper (SAP) noise and random-valued noise. 
In the presence of SAP noise, 
noisy pixels take the maximum value (i.e., 255 in 8-bit images) with a probability of $D_{s}$ or the minimum value (0) with a probability of $D_{p}$, which are respectively referred to as the salt noise (white dots) and the pepper noise (black dots) \cite{Gonzalez2018}.
The probability density function (PDF) of SAP noise in the 8-bit gray-scale images is given by \cite{Gonzalez2018}:
\begin{equation}
\label{eq:SAP}
p(z) =
\begin{cases}
D_{s} & \text{for $z=255$}\\
D_{p} & \text{for $z=0$}\\
1-D & \text{for $z=\nu$}
\end{cases}
\end{equation}
where $z$ represents intensity of the image and $\nu$ is any integer value such that $0<\nu<255$. $D_{s}$ and $D_{p}$ denote the probability of pixel corruption by salt noise and pepper noise, and $D=D_{s}+D_{p}$ is the noise density. Usually, $D_{s}$ and $D_{p}$ are considered equal.

Many attempts have been hitherto  suggested for SAP denoising (\cite{Astola1997, Esakkirajan2011, Zhang2014, Varatharajan2018, Erkan2018, Vasanth2020, Karthik2021, ROY2016Multiclass, GonzalezHidalgo2018}).
Attributed to the fact that corrupted pixels by SAP noise are the maximum or minimum values of noisy images, median-based filter approaches have received considerable attention as the starting point of the efforts to remove SAP noise. The classic median filter (MF) \cite{Bovik2005}, adaptive median filter (AMF) \cite{Hwang1995}, decision-based algorithm (DBA) \cite{Srinivasan2007}, modified decision based unsymmetric trimmed median filter (MDBUTMF) \cite{Esakkirajan2011} can be mentioned as the most famous traditional median-based filter methods. These primitive methods generally encounter formidable obstacles to eliminating SAP noise, particularly at high noise densities in which almost all the pixels in the local window are noisy \cite{Bhadouria2014, Astola1997, Zhang2014}. They also suffer from blurring the details and edges of the images and producing streaking effects \cite{Jayaraj2010, Bovik1987}.

Several algorithms using novel interpolation techniques have been recently developed with the aim of alleviating these restrictions by image information preservation and visual quality improvement, particularly in the presence of high-density noises.
Some approaches use mean-based filters on non-noisy (clean) pixels in a local window to restore noisy pixels.
Adaptive weighted mean filter (AWMF) \cite{Zhang2014} and its modified versions---e.g., adaptive switching weight mean filter (ASWMF) \cite{Thanh2020switching} and improved adaptive weighted mean filter (IAWMF) \cite{Erkan2020}---restore noisy pixels by using various weighting methods to calculate the weighted average of non-noisy pixels in an adaptive window. Riesz mean is another average-based technique that has met with success in SAP noise removal. This technique is employed in the pixel similarity-based adaptive Riesz mean filter (ARmF) \cite{Enginoglu2019} and the different adaptive modified Riesz mean filter (DAMRmF) \cite{Memis2021}.
There are also some approaches that adopt mean-based or median-based filters or a combination of both in multiple stages to denoise images. The different applied median filter (DAMF) uses two consecutive median filters. In contrast to DAMF, adaptive Cesáro mean filter (ACmF) \cite{Enginoglu2020} employs the Cesáro mean instead of the median in a recursive algorithm which improves performance but decreases the execution time. In \cite{Garg2020}, a four-stage median-average (FoMA) algorithm is introduced, employing four-step median and mean filters.

Kriging interpolation is another interpolation technique utilized for SAP noise reduction. The adaptive decision based Kriging interpolation filter (ADKIF) \cite{Varatharajan2018} can achieve good results by employing Kriging interpolation. 
Besides the aforementioned methods, a multistage selective convolution filter (MSCF) was recently proposed in \cite{Rafiee2022}, which could significantly reduce the computation time along with a considerable noise suppression performance. 
Finally, an impulse denoising method based on noise accumulation and harmonic analysis techniques (NAHAT) \cite{Thanh2021accumulation} could achieve the best SAP denoising performance among non-deep learning-based algorithms by employing noise accumulation and solving a system of equations corresponding to a harmonic function.

Recent years have seen greater emphasis than heretofore on the application of machine learning methodologies, particularly deep learning, in a wide variety of issues by benefiting from advancements in hardware architecture.
Advances in deep learning-based approaches, especially by taking advantage of convolutional neural networks (CNNs) in image data, have led to a significant breakthrough in a variety of computer vision tasks, including image reconstruction \cite{TIAN2020}, image segmentation \cite{Minaee2022}, and object classification \cite{Liu2020}.
Deep learning application in Gaussian image noise reduction has also attracted a great deal of attention as it offers vastly improved performance over classical algorithms \cite{Zhang2017, Zhang2018FFDNet, Valsesia2020, Zhang2021, Zhang2021Residual}.

Although these learning-based networks are significantly effective in removing Gaussian noise, they cannot denoise well in the presence of SAP noise, specifically at high-density noises. This poor performance of networks which are designed to remove Gaussian noise in SAP noise removal can be attributed to a variety of reasons. 
Firstly, these networks modify all pixels in noisy images; however, unlike Gaussian and Poissonian noise, SAP noise does not affect all pixels. Therefore, some pixels are noise-free, which should not alter throughout denoising. Needless to say, the quantity of clean pixels is considerable at low noise densities. 
Secondly, they restore noisy pixels using all pixels, even noisy ones, in the receptive field. In the case of SAP noise, there is no correlation between the value of noisy pixels and their original values. 
This phenomenon can be problematic in the denoising performance of the networks, particularly at high noise densities in which a considerable part of pixels is noisy. Thus, since SAP noise is pure noise, it can produce an adverse effect on the SAP noise removal function of CNNs \cite{Fu2019}. 
Finally, since SAP is not an additive noise, a residual learning strategy, which is adopted in some of these networks such as DnCNN \cite{Zhang2017}, is ineffective and introduces undesirable visual artifacts \cite{Radlak2020}.

As opposed to Gaussian noise, relatively few research studies on impulse noise reduction, specifically SAP noise, use artificial neural networks (ANNs) \cite{ Fu2019,Burger2012, Lehtinen2018, Zhang2019}. 
A non-local switching filter followed by a pretrained multi-layer perception (NLSF-MLP) model was suggested for SAP noise reduction in \cite{Burger2012}. Fu et al. proposed employing a non-local switching filter as a preprocessing stage to a CNN (NLSF-CNN) to improve the performance of SAP noise reduction from corrupted images \cite{Fu2019}. Since SAP can adversely affect the CNN models, the preprocessing stage is considered to prepare noisy images for the CNN model by changing the type of noise. Hence, NLSF-MLP and NLSF-CNN are indeed an integration of conventional and learning-based strategies. Moreover, NLSF-MLP and NLSF-CNN change all pixels of the images, which causes the performance to dwindle.


In spite of the fact that the approaches outlined above can be effective in SAP noise reduction, they are still suffering from a shortage of adequate performance and details preservation, especially at very high noise densities. 
Aiming to overcome the shortcomings of the aforementioned learning-based networks  in SAP noise removal, we propose a deep convolutional neural network, namely SeConvNet, by introducing a new selective convolutional (SeConv) block. 
The proposed network can be extremely effective in SAP denoising, particularly at very high noise densities up to 95\%, and experimental results show that it outperforms its state-of-the-art counterparts by a considerable margin in both quantitative metrics of denoising performance and visual effects.

The remainder of this paper is organized as follows. The architecture of SeConvNet and the new selective convolutional (SeConv) block are introduced in section \hyperref[sec:Model]{\ref*{sec:Model}}. Section \hyperref[sec:Experiments]{\ref*{sec:Experiments}} first depicts different phases of training SeConvNet, including data employed in the training and test stages, data preprocessing, as well as network parameters and configuration. Subsequently, experimental results are reported to assess SeConvNet performance in SAP noise removal. Finally, section \hyperref[sec:Conclusion]{\ref*{sec:Conclusion}} concludes the paper.

\section{The Proposed Denoising CNN Model}
\label{sec:Model}

According to the aforementioned adverse impact of SAP noise on CNNs, we introduce a selective convolutional (SeConv) block and employ it in the first part of the proposed network model. 
These SeConv blocks indeed prepare the network's input for the subsequent conventional convolutional layers and tackle the issue of participating pure noisy pixels 
in image reconstruction. 
Before proceeding to the network architecture, we first define the concept of noisy pixels map. 
The noisy pixels map of  an arbitrary tensor
$\boldsymbol{\mathcal{A}}$
is a tensor with the same shape as $\boldsymbol{\mathcal{A}}$ defined by
\begin{equation}
\label{eq:M}
[\mathbf{M}_{\boldsymbol{\mathcal{A}}}]_{i,j,k} =
\begin{cases}
1 & \text{if $[\boldsymbol{\mathcal{A}}]_{i,j,k} = 0$}\\
0 & \text{otherwise},
\end{cases}
\end{equation}
where $[{\bf{X}}]_{i,j,k}$  denotes the element of  ${\bf{X}}$ at coordinates $(i,j,k)$.
The non-noisy pixels map of $\boldsymbol{\mathcal{A}}$, i.e., $\mathbf{\tilde{M}}_{\boldsymbol{\mathcal{A}}}$, is obtained by flipping 1s and 0s in $\mathbf{M}_{\boldsymbol{\mathcal{A}}}$. 


\subsection{The Network Architecture}
\begin{figure*}[t]
	\centering
	\includegraphics[width=\textwidth]{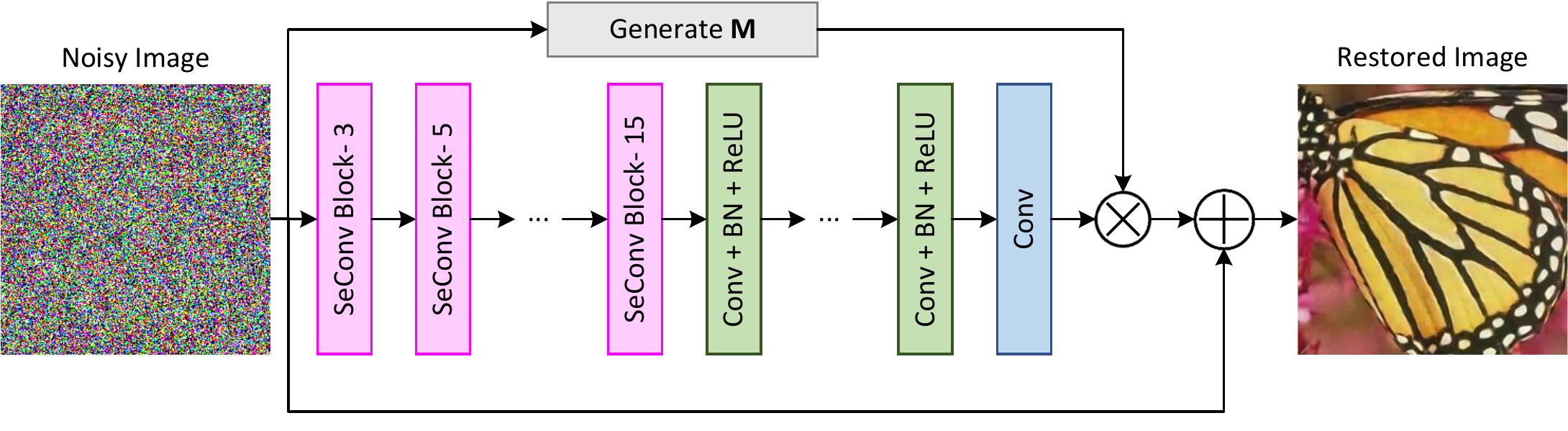}	
	\caption{The architecture of SeConvNet. Note that ``SeConv Block- $s$'' indicates a SeConv block of size $s$.}
	\label{fig:Network}
\end{figure*}
The architecture of the proposed SeConvNet model with $ D $ layers is shown in \hyperref[fig:Network]{figure~\ref*{fig:Network}}.
The input noisy image first goes through a preprocessing stage which converts pixels with a value of 255 into 0, so that all noisy pixels will be zero thereafter. As we shall see, this will simplify the processing in the following blocks.
Seven SeConv blocks constitute the first part of SeConvNet. 
SeConv blocks perform an initial SAP denoising and replace pure noisy pixels by an accurate estimate of them, whereby the subsequent convolutional layers could  improve the denoising performance.
As described in the following subsection, a SeConv block of size $ s $ selectively restores some of the noisy pixels of its input image using a trainable $ s \times s $ kernel, and non-noisy pixels remain unchanged as they pass through the SeConv blocks. 
It should also be noted that the SeConv blocks are arranged in the ascending order of $ s=3, 5, 7, 9, 11, 13, 15 $.

For the $8^{th}$ to $(D-1)^{th}$ layers, convolution (Conv) with 64 filters of size $3 \times 3$ generating 64 feature maps, along with ReLU activation are used. Furthermore, to benefit from the properties of batch normalization (BN) \cite{Zhang2017}, BN is considered between convolution and ReLU.
In order to reconstruct images, convolution with $C$ filters of size $3 \times 3$ is employed in the $D^{th}$ layer, where $C$ is the number of channels (it is equal to 1 for gray image denoising and 3 for color denoising). 

Ultimately, as it is anticipated that only noisy pixels will change, the obtained output is multiplied by the noisy pixels map of the network's input ($\mathbf{M}$), and the result is then added to the input of the network to prevent any change in clean pixels.


\subsection{Selective Convolutional Block}

As an initial denoising step, SeConv blocks are introduced to efficiently replace pure SAP noisy pixels by a first-order estimate. To provide an efficient and accurate estimation of a SAP noisy pixel, we use a normalized weighted average of non-noisy pixels in its neighborhood \cite{Zhang2014,Erkan2019}. This weighted average filtering  can be considered as a  convolution between the input image  $\mathbb{X}$ and a kernel $\omega$, in which only uncorrupted (non-noisy) pixels are used in the convolution \cite{Rafiee2022}, i.e.,
\begin{equation}
\label{eq:nonconv}
[\mathbf{S}]_{i,j,k} =
\begin{cases}
\frac{\sum\limits_{l,m,n}[\mathbb{X}]_{l,m,n}[\omega]_{i-l,j-m,k-n}}{\sum\limits_{l,m,n:[\mathbb{X}]_{l,m,n}\neq0}[\omega]_{i-l,j-m,k-n}} & \text{if ${\sum\limits_{l,m,n:[\mathbb{X}]_{m,n,k}\neq0}[\omega]_{i-l,j-m,k-n}}\neq0$}\\
0 & \text{otherwise}.
\end{cases}
\end{equation}
where the condition $l,m,n:[\mathbb{X}]_{m,n,k}\neq0$ indicates that only non-noisy pixels are incorporated in the restoration process (as all noisy pixels are turned into 0 in the preprocessing stage). Using the notion of non-noisy pixels map $\mathbf{\tilde{M}}_{\mathbb{X}}$, we can reformulate Eq. \ref{eq:nonconv} as
\begin{equation}
\label{eq:MUconv1}
[\mathbf{S}]_{i,j,k} =
\begin{cases}
\frac{\sum\limits_{l,m,n}[\mathbb{X}]_{l,m,n}[\omega]_{i-l,j-m,k-n}}{\sum\limits_{l,m,n}[\mathbf{\tilde{M}}_{\mathbb{X}}]_{l,m,n}[\omega]_{i-l,j-m,k-n}} & \text{if ${\sum\limits_{l,m,n}[\mathbf{\tilde{M}}_{\mathbb{X}}]_{l,m,n}[\omega]_{i-l,j-m,k-n}}\neq0$}\\
0 & \text{otherwise},
\end{cases}
\end{equation}
which can be equivalently restated as 
\begin{equation}
\label{eq:MUconv2}
[\mathbf{S}]_{i,j,k} = 
\begin{cases}
\frac{[\mathbb{X}\ast \omega]_{i,j,k}}{[\mathbf{\tilde{M}}_{\mathbb{X}}\ast \omega]_{i,j,k}} & \text{if $[\mathbf{\tilde{M}}_{\mathbb{X}}\ast \omega]_{i,j,k}\neq0$}\\
0 & \text{otherwise}.
\end{cases}
\end{equation}
where $  \ast $ denotes the conventional convolution operation.


For weighted average filtering to effectively restore a noisy pixel,~however, a sufficient number of non-noisy pixels should incorporate in the above convolution \cite{Varatharajan2018,Chen2020}. Accordingly, 
to further improve the initial denoising, we selectively restore only those noisy pixels for which there are at least $ \eta $ non-noisy pixels in the local $ s \times s $ around them, and the remaining noisy pixels are restored in subsequent SeConv blocks with larger kernel sizes \cite{Rafiee2022}. This will assure a minimum degree of reliability for restored pixels at each block.
It should also be noted that for effective denoising with weighted average filtering, the value of $\eta$ should be increased with the size of the kernel (since more non-noisy pixels are required in larger windows). For a SeConv block with a size of $s \times s$,  $\eta$ has been set experimentally to $s-2$.
To determine the number of non-noisy  pixels involved in the estimation of each noisy pixel, the non-noisy pixels map  $\mathbf{\tilde{M}}_{\mathbb{X}}$ is convolved with an $s \times s$ kernel of ones ($\omega_{\text{one}}$), and the reliability tensor is obtained as 
\begin{equation}
\label{eq:R}
[\mathbf{R_{\mathbb{X}}}]_{i,j,k} = 
\begin{cases}
1 & \text{if $[\mathbf{\tilde{M}}_{\mathbb{X}}\ast \omega_{\text{one}}]_{i,j,k} \geq \eta$}\\
0 & \text{otherwise.}
\end{cases}
\end{equation}
Therefore, the non-zero entries of the element-wise product of   $\mathbf{S}$, $\mathbf{M}_{\mathbb{X}}$, and $\mathbf{R}_{\mathbb{X}}$ yield the restored pixels which are qualified by the above reliability criterion. Noting thay all noisy pixels of the input were set to zero, the restored image at the output of the SeConv block is obtained as
\begin{equation}
\label{eq:SeConv}
\hat{\mathbb{X}} =  \mathbb{X} + \mathbf{S}\odot\mathbf{M}_{\mathbb{X}}\odot \mathbf{R}_{\mathbb{X}} .
\end{equation}
where $ \odot $ denotes the Hadamard (element-wise) product. 

The pseudo-code of the proposed SeConv block is shown in \hyperref[alg:SeConv Block]{algorithm \ref*{alg:SeConv Block}}.

\begin{algorithm}[t!]
	\DontPrintSemicolon
	\caption{Proposed SeConv Block.}\label{alg:SeConv Block}
	\KwIn{$\mathbb{X}, \omega, \eta$ \Comment*[r]{Input Image, Kernel, Minimum Reliability}}
	\KwOut{$\hat{\mathbb{X}}$\Comment*[r]{Output Image}} 
	
	$H, W \gets$ Size of $\mathbb{X}$\\
	$C \gets$ Number of Channels for $\mathbb{X}$ \\
	\For{$i \gets 1$ \KwTo $H$}{
		\For{$j \gets 1$ \KwTo $W$}{
			\For{$k \gets 1$ \KwTo $C$}{
				
				\eIf{$[\mathbb{X}]_{i,j,k} = 0$}{
					$[\mathbf{M}_{\mathbb{X}}]_{i,j,k} \gets 1$\\
					$[\mathbf{\tilde{M}}_{\mathbb{X}}]_{i,j,k} \gets 0$}
				{$[\mathbf{M}_{\mathbb{X}}]_{i,j,k} \gets 0$\\
					$[\mathbf{\tilde{M}}_{\mathbb{X}}]_{i,j,k} \gets 1$}
				
				\eIf{$[\mathbf{\tilde{M}}_{\mathbb{X}}\ast \omega]_{i,j,k}\neq0$}{
					$[\mathbf{S}]_{i,j,k} \gets \frac{[\mathbb{X}\ast \omega]_{i,j,k}}{[\mathbf{\tilde{M}}_{\mathbb{X}}\ast \omega]_{i,j,k}}$}
				{$[\mathbf{S}]_{i,j,k} \gets 0$}
				
				\eIf{$[\mathbf{\tilde{M}}_{\mathbb{X}}\ast \omega]_{i,j,k} \geq \eta$}{
					$[\mathbf{R_{\mathbb{X}}}]_{i,j,k} \gets 1$}
				{$[\mathbf{R_{\mathbb{X}}}]_{i,j,k} \gets 0$}
			}
		}
	}
	
	$\hat{\mathbb{X}} \gets \mathbb{X} + \mathbf{M}_{\mathbb{X}} \mathbf{R}_{\mathbb{X}} \mathbf{S}$
\end{algorithm}

\section{Experiments}
\label{sec:Experiments}
\subsection{Datasets}\label{subsec:Datasets}
\begin{figure*}[t]
	\centering
	\subfloat{
		\includegraphics[width=0.05\textwidth, height=0.05\textwidth]{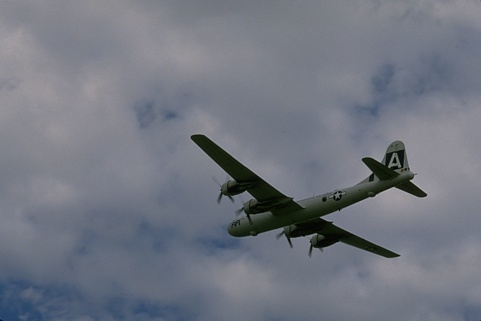}} 	\hspace*{-0.3em}
	\subfloat{
		\includegraphics[width=0.05\textwidth, height=0.05\textwidth]{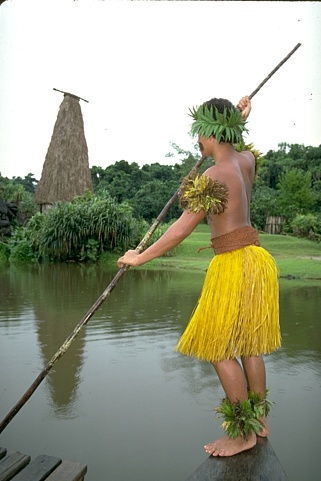}} 	\hspace*{-0.3em}
	\subfloat{
		\includegraphics[width=0.05\textwidth, height=0.05\textwidth]{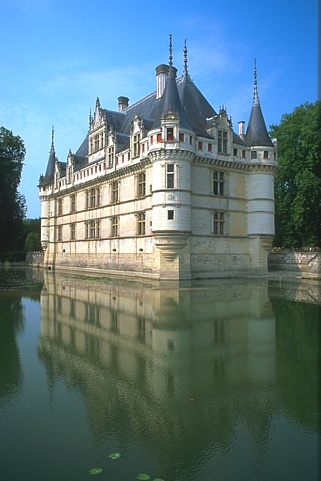}} 	\hspace*{-0.3em}
	\subfloat{
		\includegraphics[width=0.05\textwidth, height=0.05\textwidth]{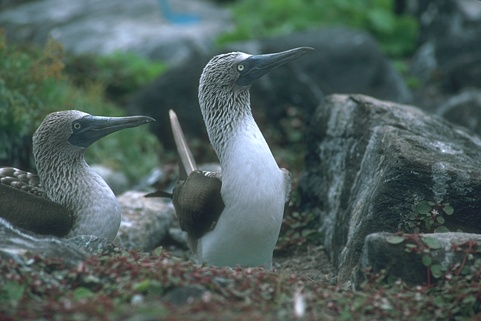}} 	\hspace*{-0.3em}
	\subfloat{
		\includegraphics[width=0.05\textwidth, height=0.05\textwidth]{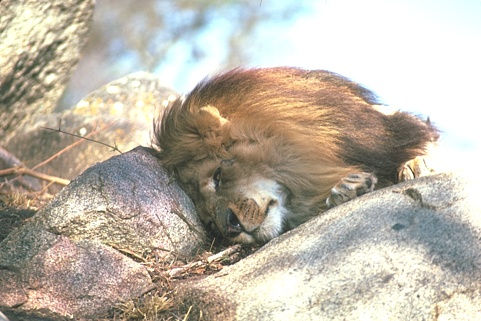}} 	\hspace*{-0.3em}	
	\subfloat{
		\includegraphics[width=0.05\textwidth, height=0.05\textwidth]{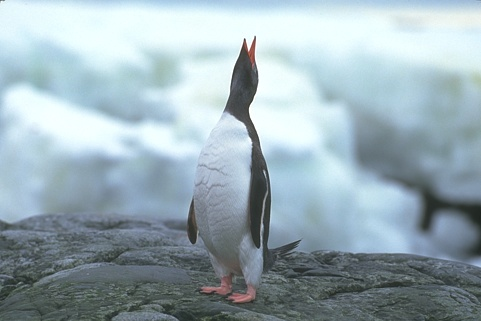}} 	\hspace*{-0.3em}
	\subfloat{
		\includegraphics[width=0.05\textwidth, height=0.05\textwidth]{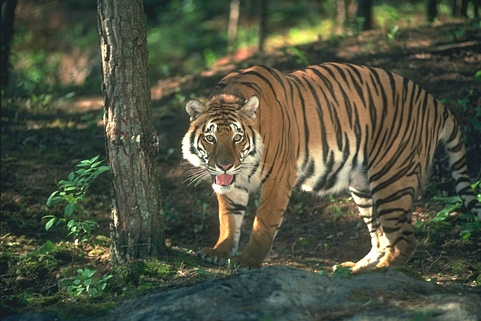}} 	\hspace*{-0.3em}
	\subfloat{
		\includegraphics[width=0.05\textwidth, height=0.05\textwidth]{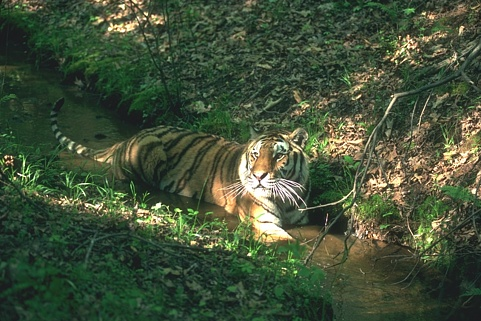}} 	\hspace*{-0.3em}
	\subfloat{
		\includegraphics[width=0.05\textwidth, height=0.05\textwidth]{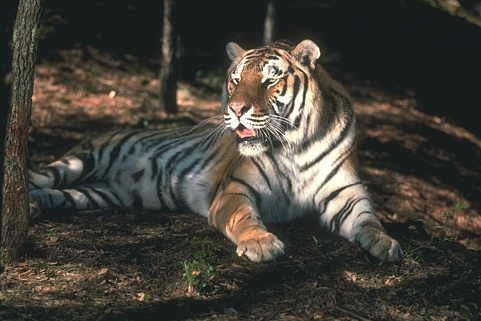}} 	\hspace*{-0.3em}
	\subfloat{
		\includegraphics[width=0.05\textwidth, height=0.05\textwidth]{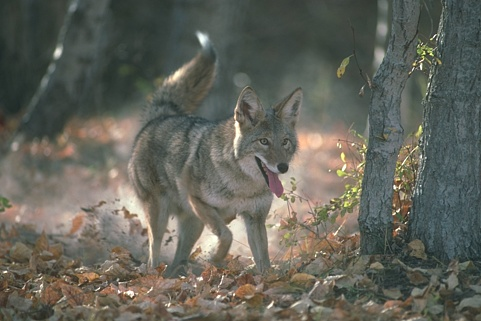}} 	\hspace*{-0.3em}	
	\subfloat{
		\includegraphics[width=0.05\textwidth, height=0.05\textwidth]{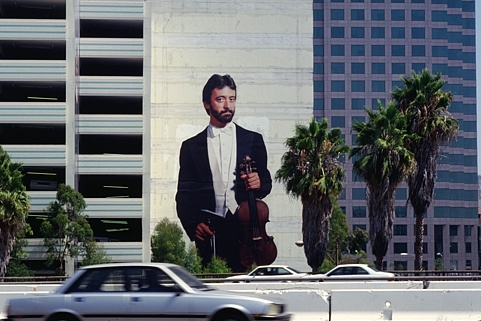}} 	\hspace*{-0.3em}	
	\subfloat{
		\includegraphics[width=0.05\textwidth, height=0.05\textwidth]{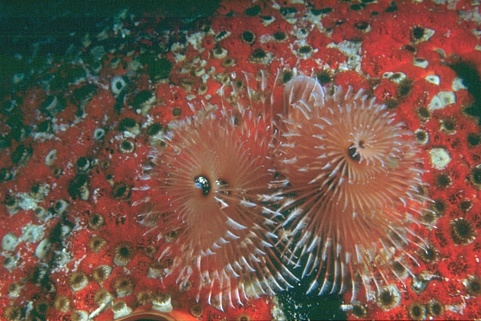}} 	\hspace*{-0.3em}	
	\subfloat{
		\includegraphics[width=0.05\textwidth, height=0.05\textwidth]{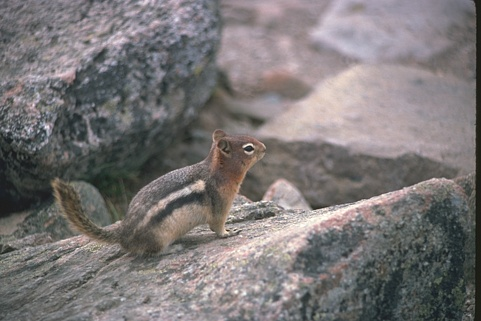}} 	\hspace*{-0.3em}	
	\subfloat{
		\includegraphics[width=0.05\textwidth, height=0.05\textwidth]{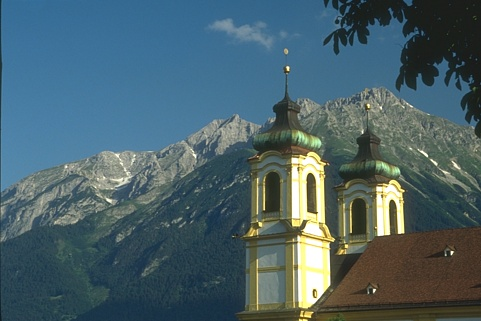}} 	\hspace*{-0.3em}	
	\subfloat{
		\includegraphics[width=0.05\textwidth, height=0.05\textwidth]{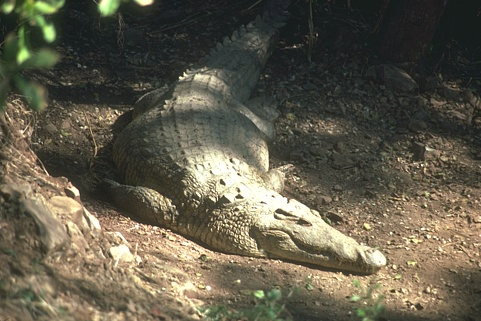}} 	\hspace*{-0.3em}	
	\subfloat{
		\includegraphics[width=0.05\textwidth, height=0.05\textwidth]{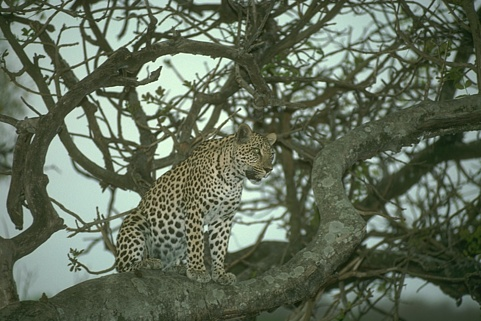}} 	\hspace*{-0.3em}	
	\subfloat{
		\includegraphics[width=0.05\textwidth, height=0.05\textwidth]{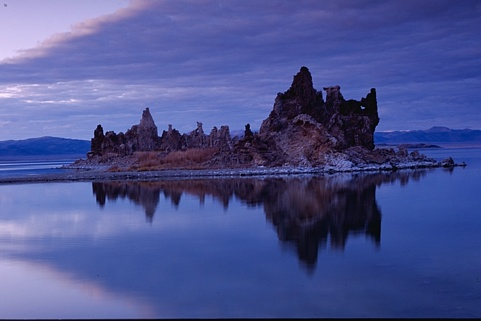}}
	\\[-1.5ex]
	\subfloat{
		\includegraphics[width=0.05\textwidth, height=0.05\textwidth]{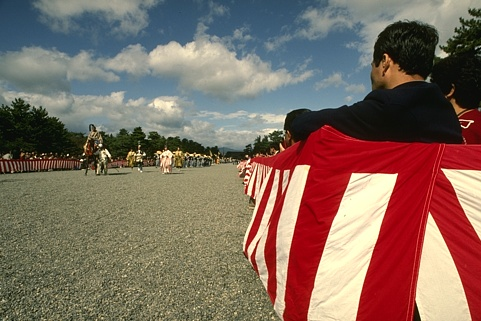}} 	\hspace*{-0.3em}
	\subfloat{
		\includegraphics[width=0.05\textwidth, height=0.05\textwidth]{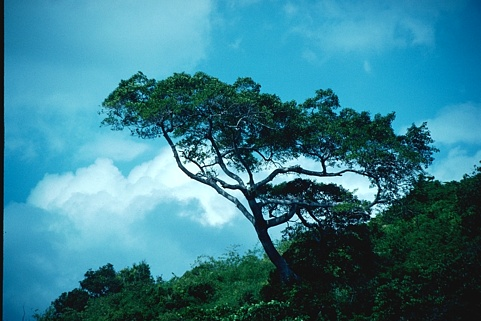}} 	\hspace*{-0.3em}
	\subfloat{
		\includegraphics[width=0.05\textwidth, height=0.05\textwidth]{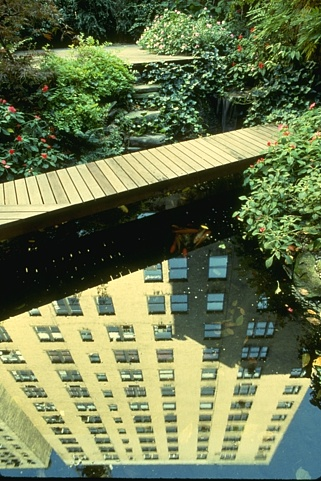}} 	\hspace*{-0.3em}	
	\subfloat{
		\includegraphics[width=0.05\textwidth, height=0.05\textwidth]{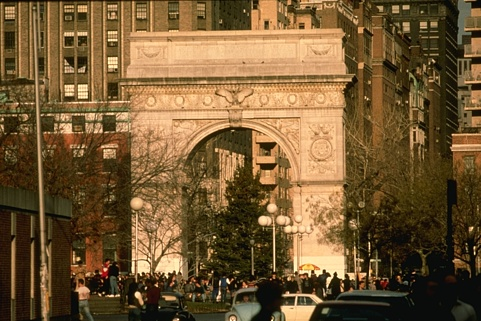}} 	\hspace*{-0.3em}	
	\subfloat{
		\includegraphics[width=0.05\textwidth, height=0.05\textwidth]{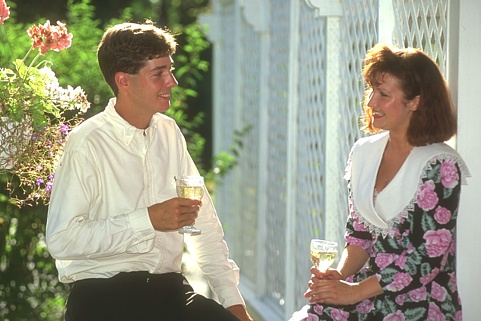}} 	\hspace*{-0.3em}	
	\subfloat{
		\includegraphics[width=0.05\textwidth, height=0.05\textwidth]{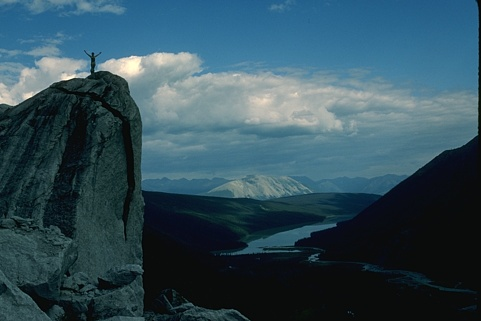}} 	\hspace*{-0.3em}	
	\subfloat{
		\includegraphics[width=0.05\textwidth, height=0.05\textwidth]{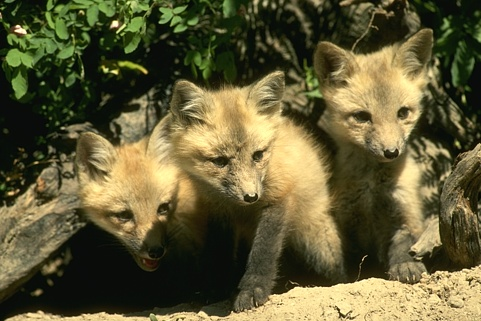}} 	\hspace*{-0.3em}	
	\subfloat{
		\includegraphics[width=0.05\textwidth, height=0.05\textwidth]{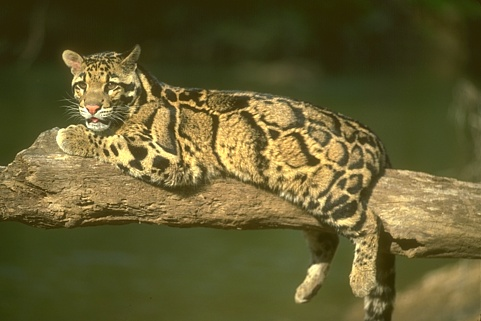}} 	\hspace*{-0.3em}	
	\subfloat{
		\includegraphics[width=0.05\textwidth, height=0.05\textwidth]{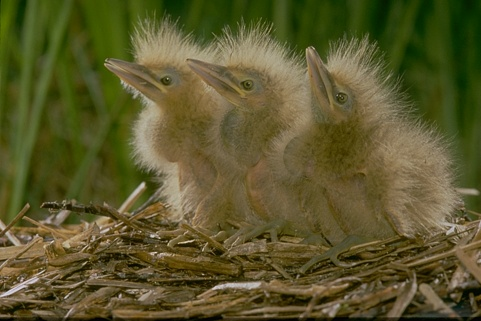}} 	\hspace*{-0.3em}	
	\subfloat{
		\includegraphics[width=0.05\textwidth, height=0.05\textwidth]{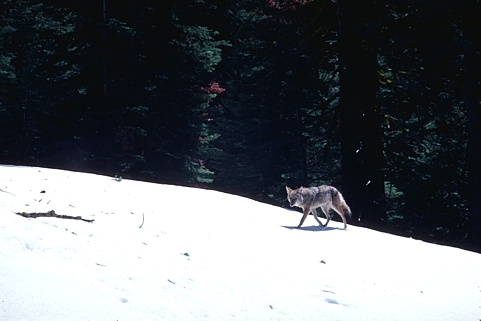}} 	\hspace*{-0.3em}
	\subfloat{
		\includegraphics[width=0.05\textwidth, height=0.05\textwidth]{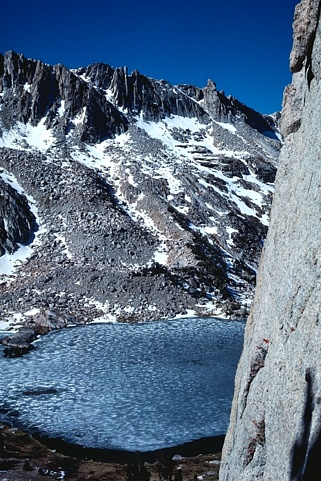}} 	\hspace*{-0.3em}
	\subfloat{
		\includegraphics[width=0.05\textwidth, height=0.05\textwidth]{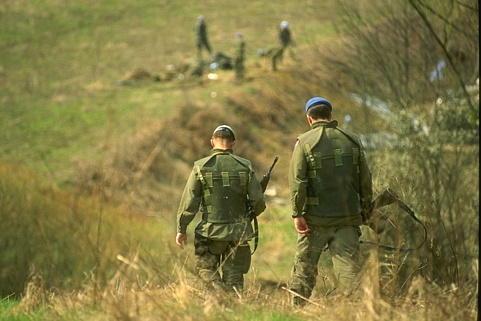}} 	\hspace*{-0.3em}
	\subfloat{
		\includegraphics[width=0.05\textwidth, height=0.05\textwidth]{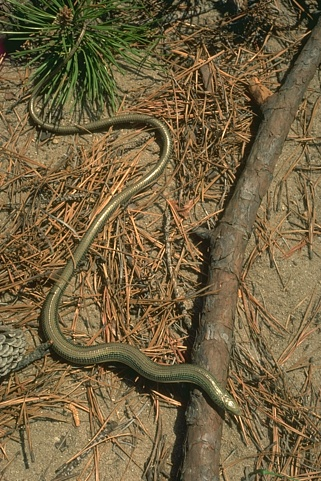}} 	\hspace*{-0.3em}	
	\subfloat{
		\includegraphics[width=0.05\textwidth, height=0.05\textwidth]{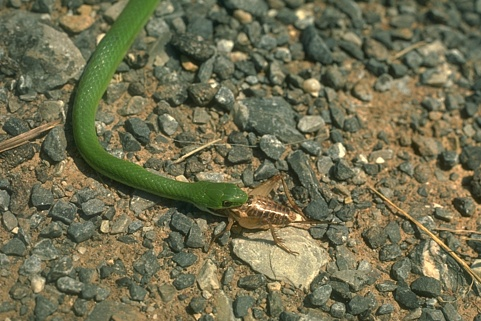}} 	\hspace*{-0.3em}	
	\subfloat{
		\includegraphics[width=0.05\textwidth, height=0.05\textwidth]{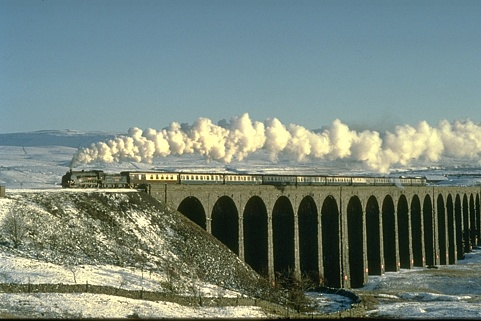}} 	\hspace*{-0.3em}	
	\subfloat{
		\includegraphics[width=0.05\textwidth, height=0.05\textwidth]{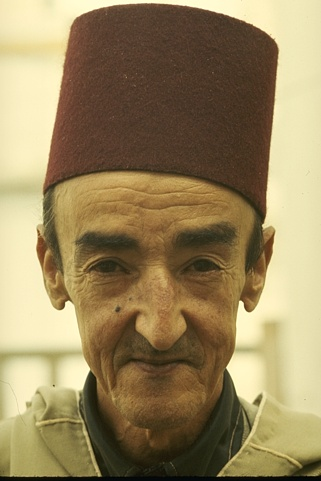}} 	\hspace*{-0.3em}	
	\subfloat{
		\includegraphics[width=0.05\textwidth, height=0.05\textwidth]{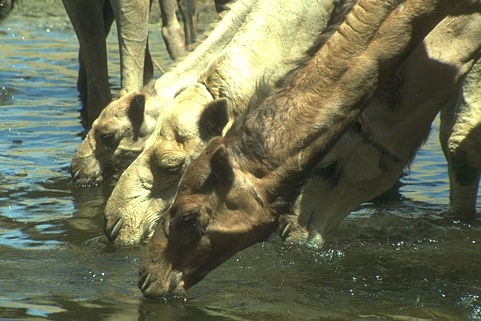}} 	
	\\[-1.5ex]
	\subfloat{
		\includegraphics[width=0.05\textwidth, height=0.05\textwidth]{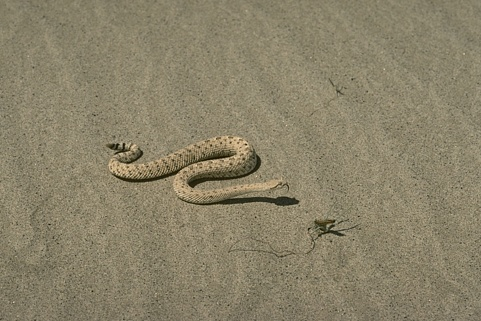}} 	\hspace*{-0.3em}	
	\subfloat{
		\includegraphics[width=0.05\textwidth, height=0.05\textwidth]{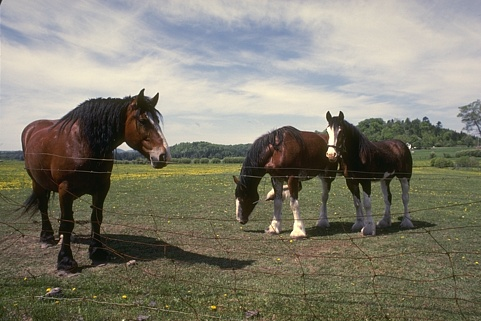}} 	\hspace*{-0.3em}	
	\subfloat{
		\includegraphics[width=0.05\textwidth, height=0.05\textwidth]{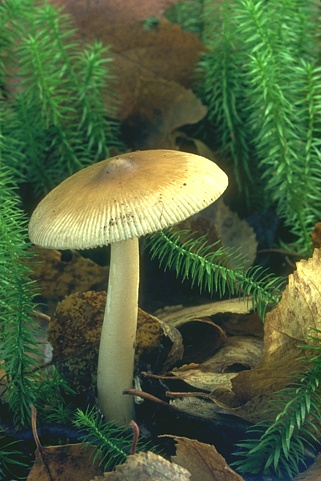}} 	\hspace*{-0.3em}
	\subfloat{
		\includegraphics[width=0.05\textwidth, height=0.05\textwidth]{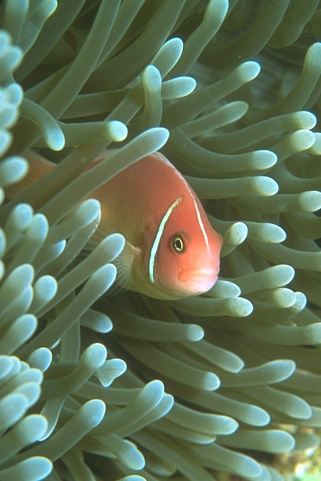}} 	\hspace*{-0.3em}
	\subfloat{
		\includegraphics[width=0.05\textwidth, height=0.05\textwidth]{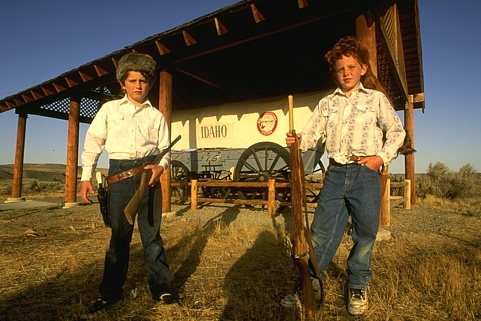}} 	\hspace*{-0.3em}
	\subfloat{
		\includegraphics[width=0.05\textwidth, height=0.05\textwidth]{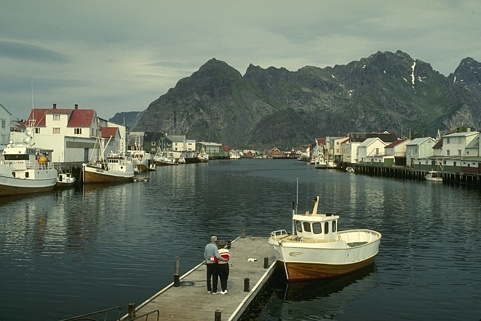}} 	\hspace*{-0.3em}	
	\subfloat{
		\includegraphics[width=0.05\textwidth, height=0.05\textwidth]{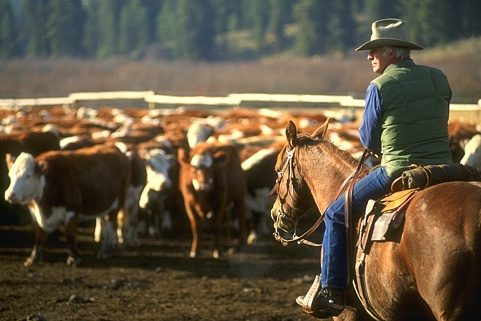}} 	\hspace*{-0.3em}	
	\subfloat{
		\includegraphics[width=0.05\textwidth, height=0.05\textwidth]{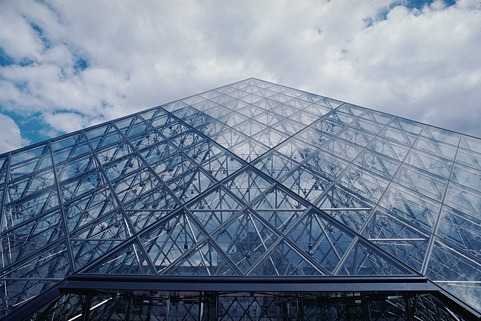}} 	\hspace*{-0.3em}	
	\subfloat{
		\includegraphics[width=0.05\textwidth, height=0.05\textwidth]{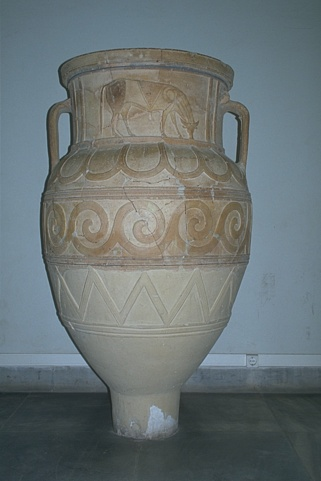}} 	\hspace*{-0.3em}	
	\subfloat{
		\includegraphics[width=0.05\textwidth, height=0.05\textwidth]{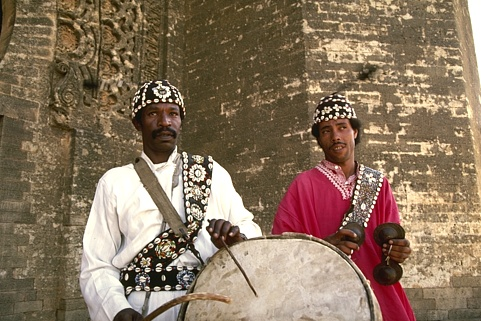}} 	\hspace*{-0.3em}	
	\subfloat{
		\includegraphics[width=0.05\textwidth, height=0.05\textwidth]{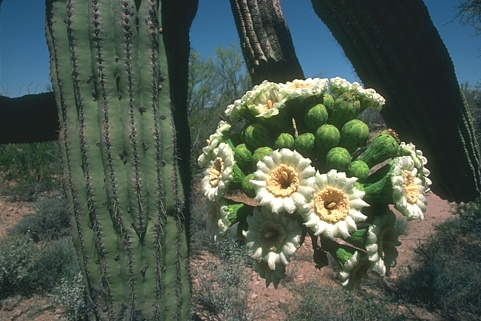}} 	\hspace*{-0.3em}	
	\subfloat{
		\includegraphics[width=0.05\textwidth, height=0.05\textwidth]{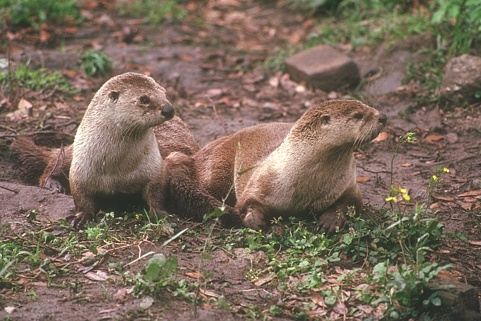}} 	\hspace*{-0.3em}	
	\subfloat{
		\includegraphics[width=0.05\textwidth, height=0.05\textwidth]{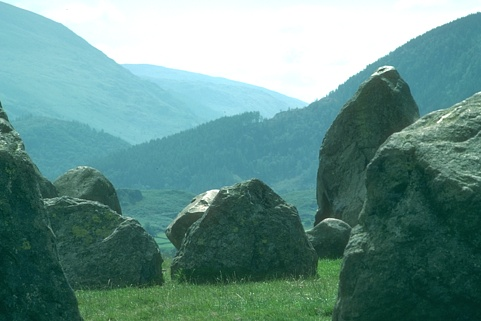}} 	\hspace*{-0.3em}
	\subfloat{
		\includegraphics[width=0.05\textwidth, height=0.05\textwidth]{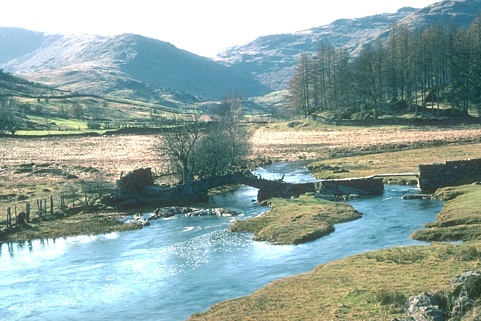}} 	\hspace*{-0.3em}
	\subfloat{
		\includegraphics[width=0.05\textwidth, height=0.05\textwidth]{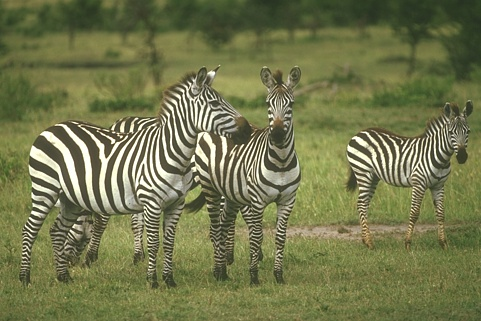}} 	\hspace*{-0.3em}
	\subfloat{
		\includegraphics[width=0.05\textwidth, height=0.05\textwidth]{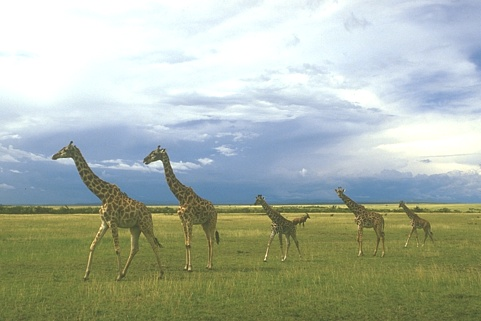}} 	\hspace*{-0.3em}	
	\subfloat{
		\includegraphics[width=0.05\textwidth, height=0.05\textwidth]{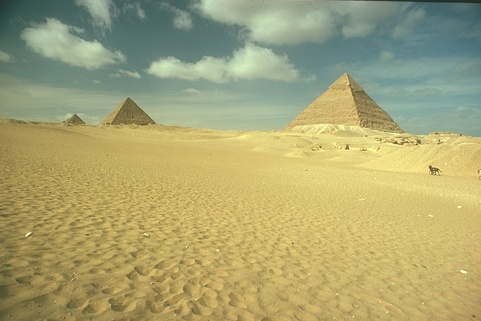}} 
	\\[-1.5ex]
	\subfloat{
		\includegraphics[width=0.05\textwidth, height=0.05\textwidth]{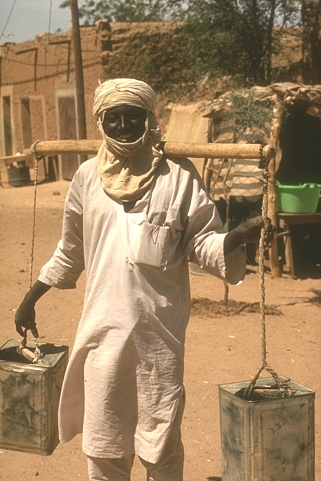}} 	\hspace*{-0.3em}	
	\subfloat{
		\includegraphics[width=0.05\textwidth, height=0.05\textwidth]{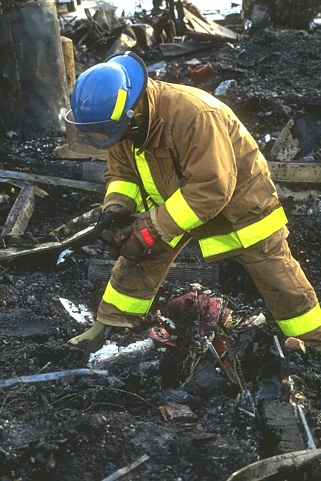}} 	\hspace*{-0.3em}	
	\subfloat{
		\includegraphics[width=0.05\textwidth, height=0.05\textwidth]{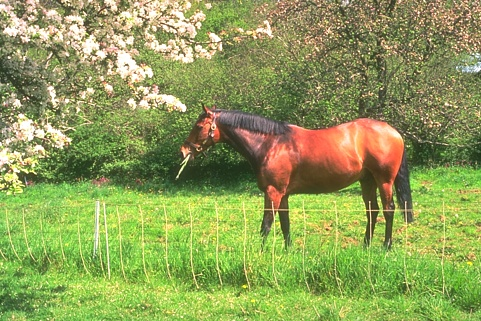}} 	\hspace*{-0.3em}	
	\subfloat{
		\includegraphics[width=0.05\textwidth, height=0.05\textwidth]{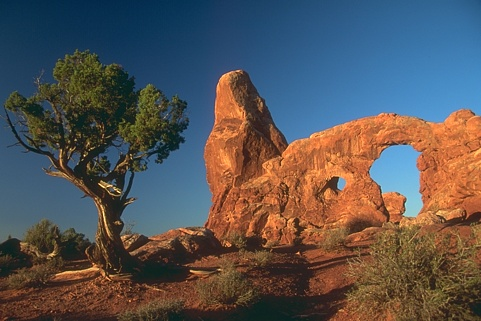}} 	\hspace*{-0.3em}	
	\subfloat{
		\includegraphics[width=0.05\textwidth, height=0.05\textwidth]{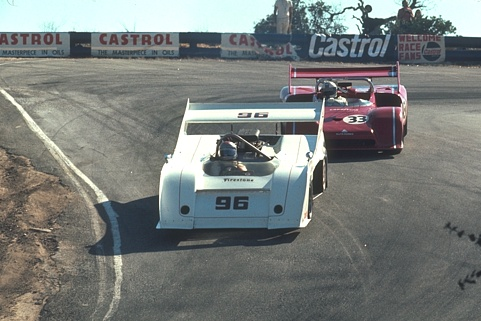}} 	\hspace*{-0.3em}	
	\subfloat{
		\includegraphics[width=0.05\textwidth, height=0.05\textwidth]{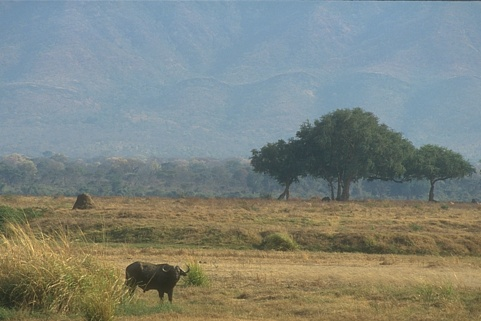}} 	\hspace*{-0.3em}
	\subfloat{
		\includegraphics[width=0.05\textwidth, height=0.05\textwidth]{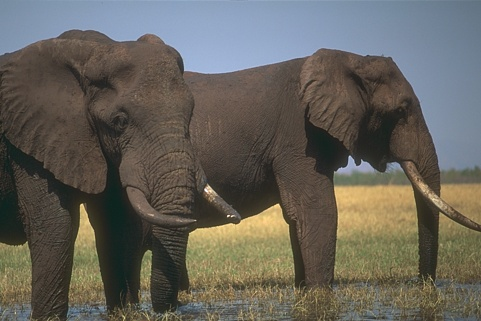}} 	\hspace*{-0.3em}
	\subfloat{
		\includegraphics[width=0.05\textwidth, height=0.05\textwidth]{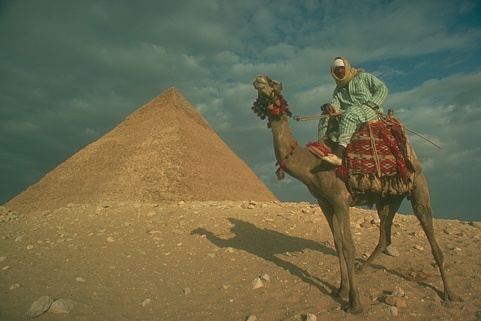}} 	\hspace*{-0.3em}
	\subfloat{
		\includegraphics[width=0.05\textwidth, height=0.05\textwidth]{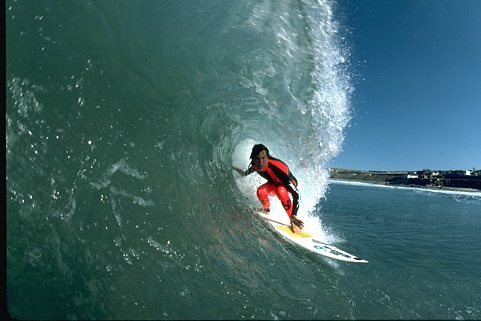}} 	\hspace*{-0.3em}	
	\subfloat{
		\includegraphics[width=0.05\textwidth, height=0.05\textwidth]{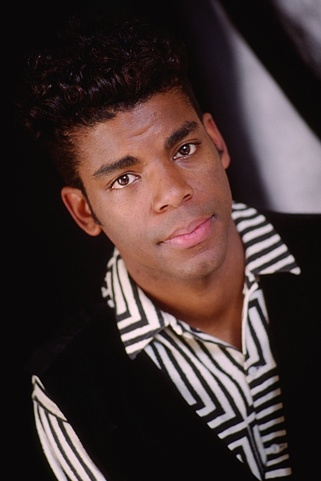}} 	\hspace*{-0.3em}	
	\subfloat{
		\includegraphics[width=0.05\textwidth, height=0.05\textwidth]{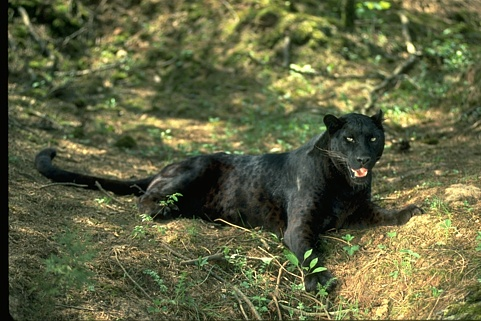}} 	\hspace*{-0.3em}	
	\subfloat{
		\includegraphics[width=0.05\textwidth, height=0.05\textwidth]{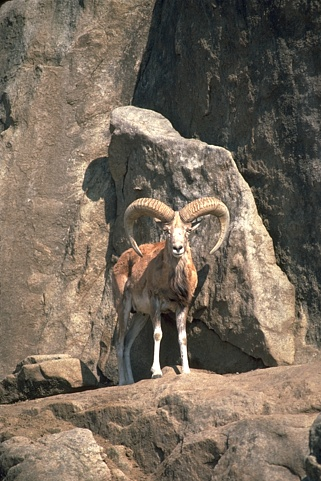}} 	\hspace*{-0.3em}	
	\subfloat{
		\includegraphics[width=0.05\textwidth, height=0.05\textwidth]{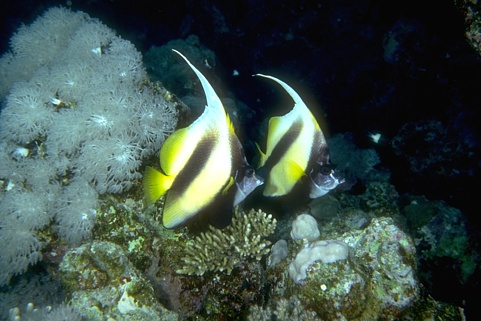}} 	\hspace*{-0.3em}	
	\subfloat{
		\includegraphics[width=0.05\textwidth, height=0.05\textwidth]{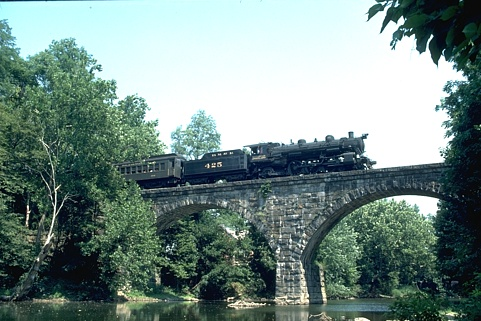}} 	\hspace*{-0.3em}	
	\subfloat{
		\includegraphics[width=0.05\textwidth, height=0.05\textwidth]{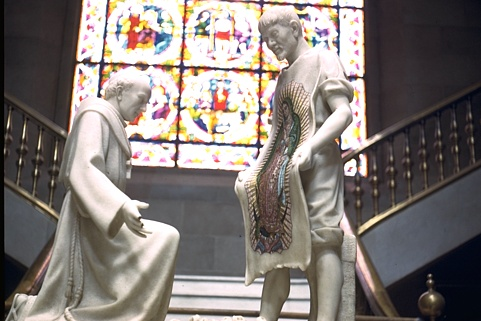}} 	\hspace*{-0.3em}	
	\subfloat{
		\includegraphics[width=0.05\textwidth, height=0.05\textwidth]{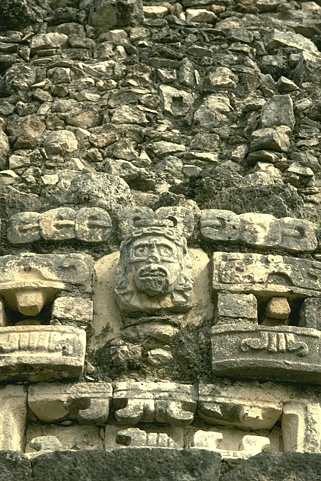}} 	\hspace*{-0.3em}
	\subfloat{
		\includegraphics[width=0.05\textwidth, height=0.05\textwidth]{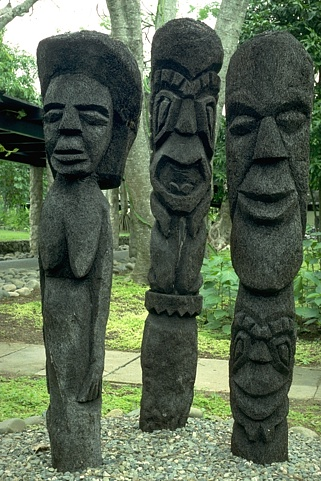}} 
	
	\caption{The BSD68 dataset.}
	\label{fig:BSD68}
	\hspace*{-0.9em}%
\end{figure*}

\begin{figure*}[t]
	\centering
	\subfloat[]{\label{fig:TI20:baboon}
		\includegraphics[width=0.09\textwidth]{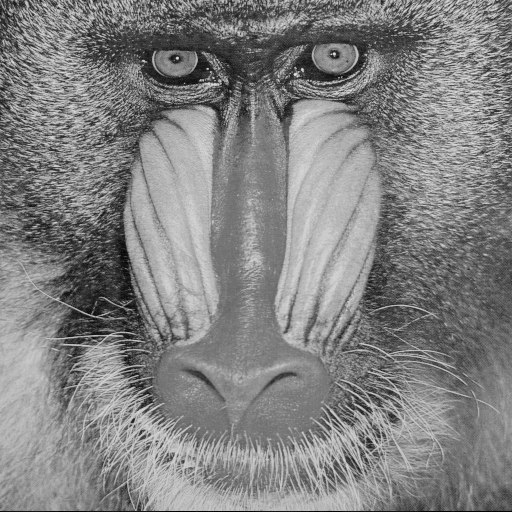}}
	\subfloat[]{\label{fig:TI20:barbara}
		\includegraphics[width=0.09\textwidth]{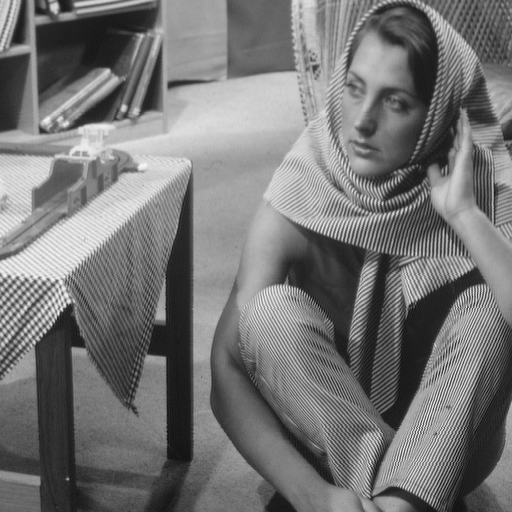}}
	\subfloat[]{\label{fig:TI20:blonde_woman}
		\includegraphics[width=0.09\textwidth]{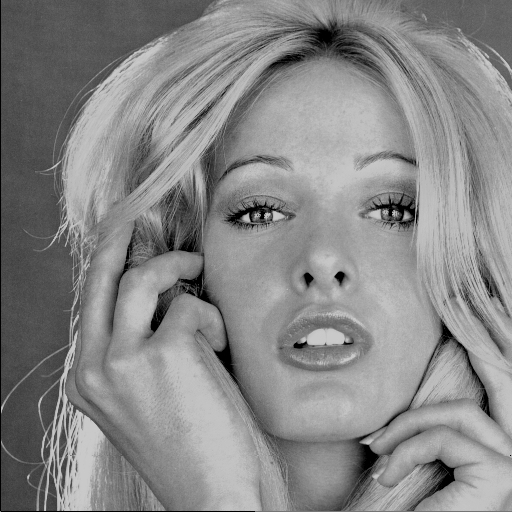}}
	\subfloat[]{\label{fig:TI20:boat}
		\includegraphics[width=0.09\textwidth]{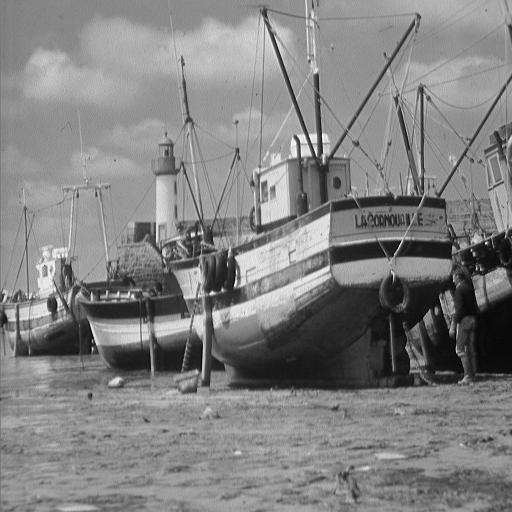}}	
	\subfloat[]{\label{fig:TI20:bridge}
		\includegraphics[width=0.09\textwidth]{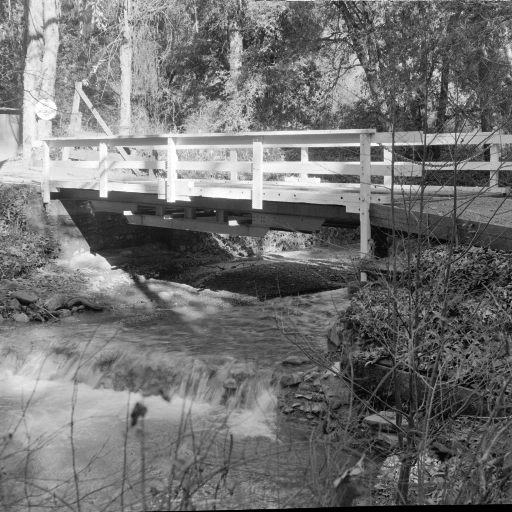}}	
	\subfloat[]{\label{fig:TI20:cameraman}
		\includegraphics[width=0.09\textwidth]{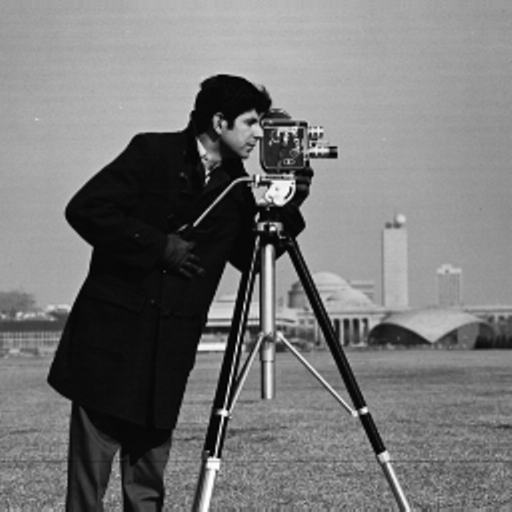}}
	\subfloat[]{\label{fig:TI20:dark_haired_woman}
		\includegraphics[width=0.09\textwidth]{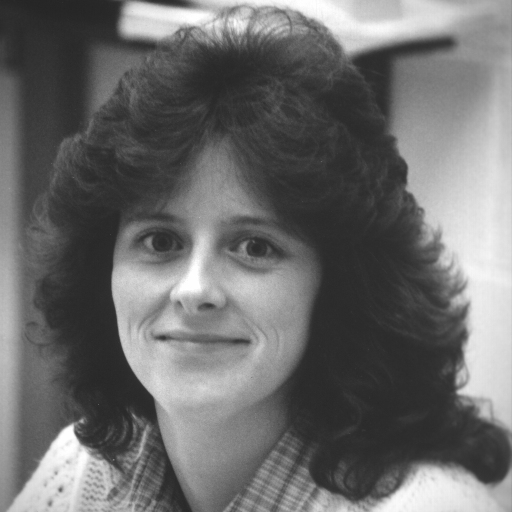}}	
	\subfloat[]{\label{fig:TI20:einstein}
		\includegraphics[width=0.09\textwidth]{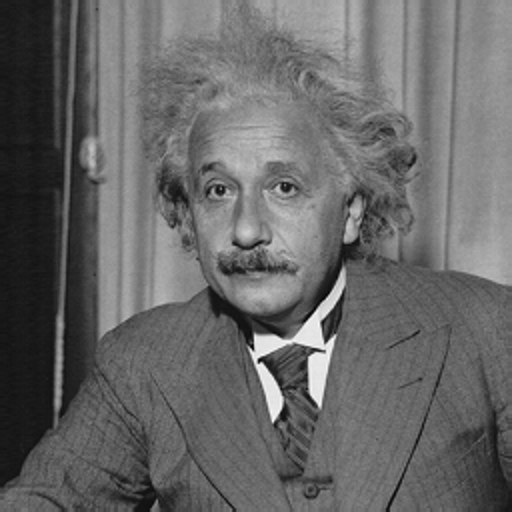}}
	\subfloat[]{\label{fig:TI20:elaine}
		\includegraphics[width=0.09\textwidth]{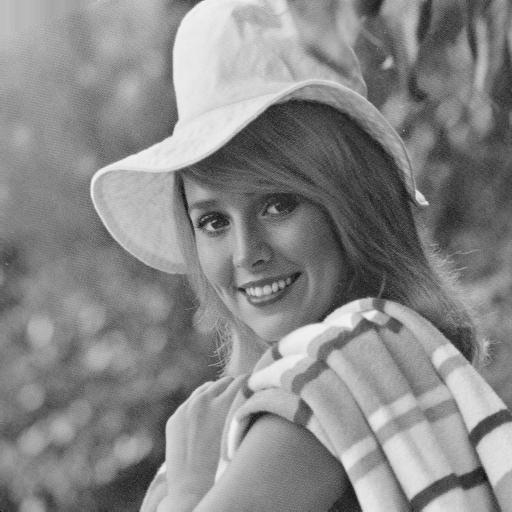}}
	\subfloat[]{\label{fig:TI20:flintstones}
		\includegraphics[width=0.09\textwidth]{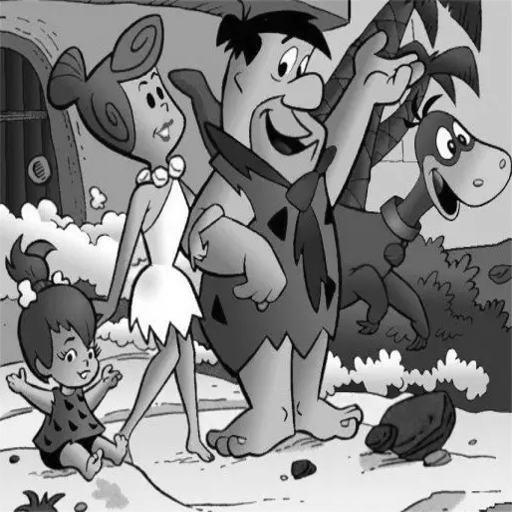}}	
	\\[-2ex]	
	\subfloat[]{\label{fig:TI20:flower}
		\includegraphics[width=0.09\textwidth]{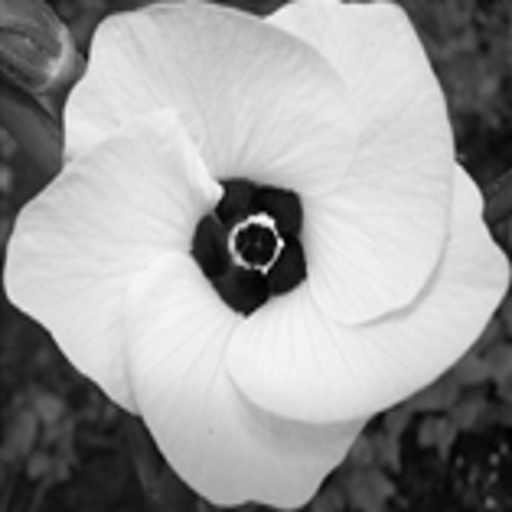}}
	\subfloat[]{\label{fig:TI20:hill}
		\includegraphics[width=0.09\textwidth]{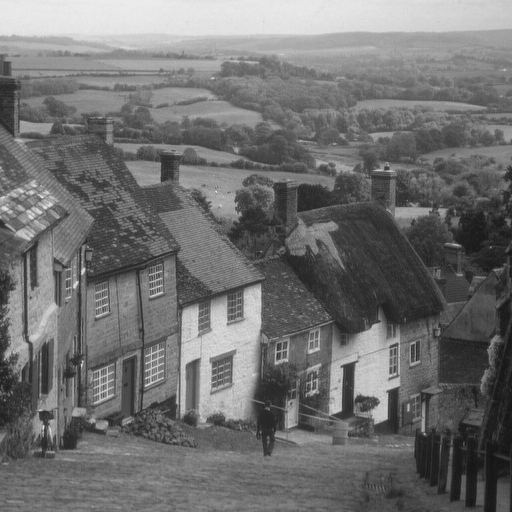}}
	\subfloat[]{\label{fig:TI20:house}
		\includegraphics[width=0.09\textwidth]{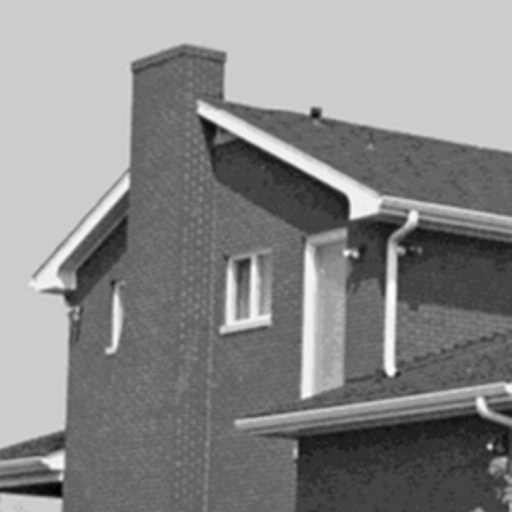}}	
	\subfloat[]{\label{fig:TI20:jetplane}
		\includegraphics[width=0.09\textwidth]{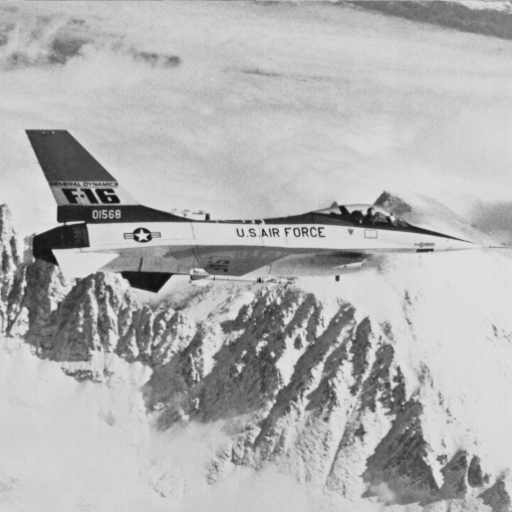}}
	\subfloat[]{\label{fig:TI20:lake}
		\includegraphics[width=0.09\textwidth]{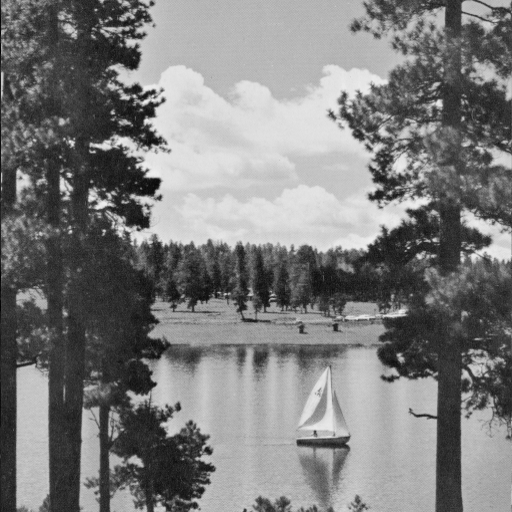}}	
	\subfloat[]{\label{fig:TI20:lena}
		\includegraphics[width=0.09\textwidth]{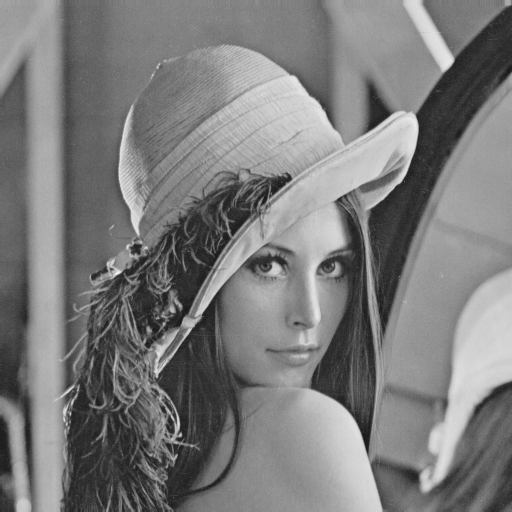}}	
	\subfloat[]{\label{fig:TI20:living_room}
		\includegraphics[width=0.09\textwidth]{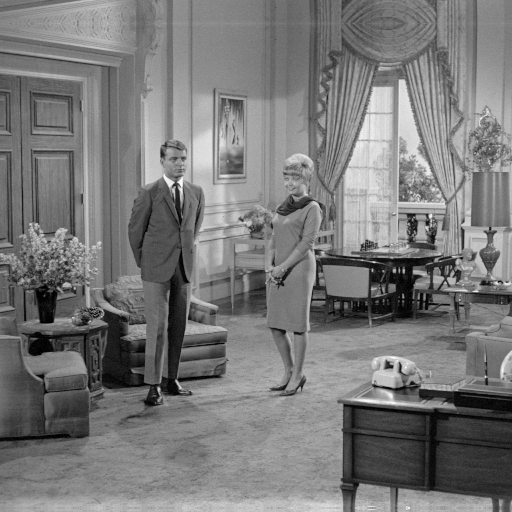}}	
	\subfloat[]{\label{fig:TI20:parrot}
		\includegraphics[width=0.09\textwidth]{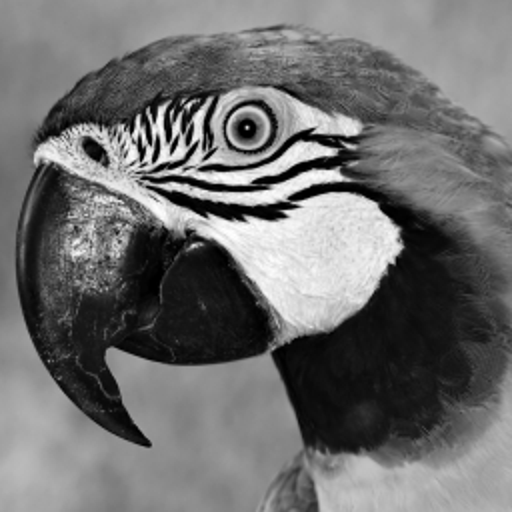}}
	\subfloat[]{\label{fig:TI20:peppers}
		\includegraphics[width=0.09\textwidth]{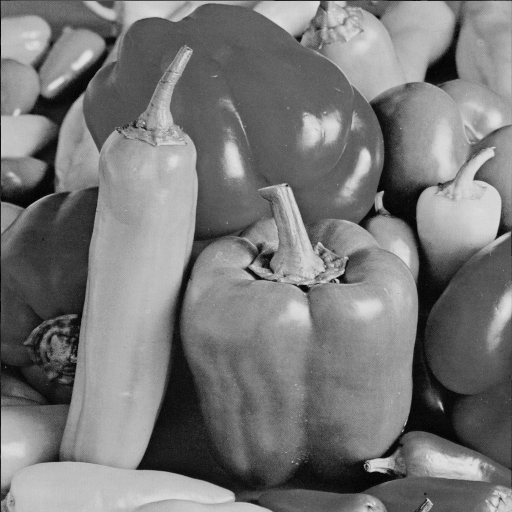}}
	\subfloat[]{\label{fig:TI20:pirate}
		\includegraphics[width=0.09\textwidth]{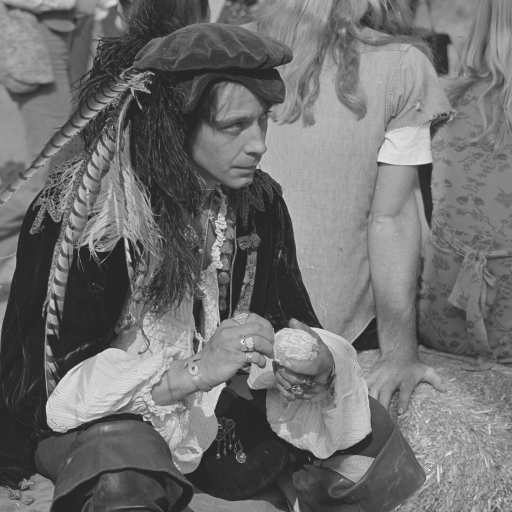}}
	
	\caption{The 20 traditional test images dataset. (a)~Baboon (b)~Barbara (c)~Blonde Woman (d)~Boat (e)~Bridge (f)~Cameraman (g)~Dark Haired Woman (h)~Einstein (i)~Elaine (j)~Flintstones (k)~Flower (l)~Hill (m)~House (n)~Jet Plane (o)~Lake (p)~Lena (q)~Living Room (r)~Parrot (s)~Peppers (t)~Pirate.}
	\label{fig:TI20}
\end{figure*}

In the training stage of SeConvNet,  following \cite{Chen2017}, we consider 400 images with sizes of $180 \times 180$ as the training dataset for {gray-scale image denoising}.
To assess the SAP noise removal performance, we employ two widely used datasets  in the test phase. The first dataset contains a gray-scale version of 68 natural images of sizes 321-by-481 and 481-by-321 (\hyperref[fig:BSD68]{figure \ref*{fig:BSD68}}), which are part of the Berkeley segmentation dataset (BSD68) \cite{Martin2001}. 
The second is the 20 traditional test images dataset. This dataset comprises the $512 \times 512$ gray-scale versions of ``Baboon'', ``Barbara'', ``Blonde Woman'', ``Boat'', ``Bridge'', ``Cameraman'', ``Dark Haired Woman'', ``Einstein'', ``Elaine'', ``Flintstones'', ``Flower'', ``Hill'', ``House'', ``Jet Plane'', ``Lake'', ``Lena'', ``Living Room'', ``Parrot'', ``Peppers'', and ``Pirate'' images shown in \hyperref[fig:TI20]{figure \ref*{fig:TI20}}.

Moreover, for color image denoising we train SeConvNet using the rest of the 500 color images in the Berkeley segmentation dataset except BSD68. 
We also utilized the color version of BSD68 (CBSD68) 
as the test data.

We choose the size of the patches for training as $40 \times 40$ by cropping the data, which is augmented by random rotation, random flip and random resizing with various scale factors of 1, 0.9, 0.8, and 0.7. We train the SeConvNet using the training dataset and its corrupted version by SAP noise with known specific noise density. 

\subsection{Preprocessing Stage}
As a preprocessing step, all noisy pixels are set to zero before feeding to SeConvNet. The process can be easily accomplished by converting all pixels with values of 255 to 0. The principal reason is to make all noisy pixels numerically identical.

\subsection{Network Training}
To meet the need for enough spatial information in the noise reduction task and also to consider the complexity of the network, we set the network depth $D$ to 27. 
We utilize ones and orthogonal matrices as the initial weights of the SeConv blocks and convolutions, respectively. Moreover, all biases are set to zero during the training stage. 
The following loss function
is adopted to train the trainable parameters of the network $\Theta$ for predicting the clean images using $P$ patches
\begin{equation}
\mathcal{L}(\Theta)=\frac{1}{2P} \sum_{i=1}^{P}\left\|\mathcal{F}\left(\mathbf{X}_{i}; \Theta\right)-\mathbf{Y}_{i}\right\|^{2}.
\end{equation}
where  $\mathbf{\hat{X}}=\mathcal{F}\left(\mathbf{X}; \Theta\right)$ is the estimated image from  the input noisy observation $\mathbf{X}$, and $\mathbf{Y}$ is the desired clean image. The loss function is minimized using the ADAM solver algorithm \cite{ADAM2015} with $\alpha$ of $1\mathit{e}{-3}$, $\beta1$ of 0.9, $\beta2$ of 0.999, and $\epsilon$ of $1\mathit{e}{-7}$. 
The network is trained 50 epochs with a batch size of 128. The learning rate declines from $1\mathit{e}{-3}$ to $1\mathit{e}{-4}$. The model was trained in the Python Programming Language using the TensorFlow library running on a 64-bit operating system with an Intel\textregistered\ Core\texttrademark\ i9-9900K processor, 64GB of RAM, and an NVIDIA GeForce RTX 2070 graphics card.

\subsection{Evaluation Criteria}
Peak signal-to-noise ratio (PSNR) is widely used to measure the performance of different denoising approaches. A higher PSNR value indicates that the denoised image, in general, is closer to the original image and has a higher visual quality. In addition, the structural similarity index measure (SSIM) criterion \cite{Zhou2004} is utilized as well, since PSNR would yield unreliable results in some cases \cite{Hore2010}. SSIM is a measurement of structural similarity between two images, and its value ranges between 0 and 1. The higher the SSIM value, the more structural similarity.

For an original (desired clean) image $\mathbf{Y}$   and its denoised image $\mathbf{\hat{X}}$ of size $H \times W$ and $C$ channels, PSNR (dB) is defined as
\begin{equation}
\label{eq:psnr}
\text{PSNR}\left(\mathbf{\hat{X}},\mathbf{Y}\right) = 10 \log_{10} \left(\frac{255^{2}}{\text{MSE}(\mathbf{\hat{X}},\mathbf{Y})}\right),
\end{equation}
where 
\begin{equation}
\label{eq:mse}
\text{MSE}\left(\mathbf{\hat{X}},\mathbf{Y}\right) = \frac{1}{CHW} \sum_{k=1}^{C} \sum_{i=1}^{H} \sum_{j=1}^{W} \left(\mathbf{\hat{X}}_{i,j,k}-\mathbf{Y}_{i,j,k}\right)^{2}
\end{equation}	
is the  mean square error (MSE) between $\mathbf{\hat{X}}$ and $\mathbf{Y}$, and SSIM is  defined as 
\begin{equation}
\label{eq:ssim}
\text{SSIM}\left(\mathbf{\hat{X}},\mathbf{Y}\right) = \frac{\left(2\mu_{\mathbf{\hat{X}}}\mu_{\mathbf{Y}}+c_{1}\right)\left(2\sigma_{\mathbf{\hat{X}}\mathbf{Y}}+c_{2}\right)}{\left(\mu_{\mathbf{\hat{X}}}^{2}+\mu_{\mathbf{Y}}^{2}+c_{1}\right)\left(\sigma_{\mathbf{\hat{X}}}^{2}+\sigma_{\mathbf{Y}}^{2}+c_{2}\right)}.
\end{equation}
Here $\mu_{\mathbf{\hat{X}}}$, $\mu_{\mathbf{Y}}$, $\sigma_{\mathbf{\hat{X}}}$, $\sigma_{\mathbf{Y}}$, and $\sigma_{\mathbf{\hat{X}}\mathbf{Y}}$ indicate, respectively, the average intensity of $\mathbf{\hat{X}}$, average intensity of $\mathbf{Y}$, standard deviation of $\mathbf{\hat{X}}$, standard deviation of $\mathbf{Y}$, and cross-covariance of $\mathbf{\hat{X}}$ and $\mathbf{Y}$. The two small constants $c_{1}=(k_{1}L)^{2}$ and $c_{2}=(k_{2}L)^{2}$ are used for division stabilization. The value of $L$ is assumed to be 255 for 8-bit gray-scale version of images and the default values of $k_{1}$, $k_{2}$ are considered to be 0.01 and 0.03.

\subsection{Experiments on SAP noise Removal}
In this subsection, we discuss the results of SAP denoising measured by PSNR and SSIM.
A comparison is provided between the proposed SeConvNet and seven state-of-the-art methods, including ADKIF \cite{Varatharajan2018}, NLSF-MLP \cite{Burger2012}, NLSF-CNN \cite{Fu2019}, ARmF \cite{Enginoglu2019}, ACmF \cite{Enginoglu2020}, IAWMF \cite{Erkan2020}, NAHAT \cite{Thanh2021accumulation}, and DAMRmF \cite{Memis2021}. 
The source code of SeConvNet is available at \url{https://github.com/AliRafiee7/SeConvNet}.

{\renewcommand{\arraystretch}{1.2}
\begin{table*}[t]
	\caption{Results of PSNR (dB) for different methods at different noise densities on the 20 traditional test images and BSD68 gray-scale datasets.}
	\label{table:psnr:gray}
	\centering
	\resizebox{\textwidth}{!}{
	\begin{tabular}{clrrrrrrrrrrr}
		\hline 
		Dataset & Methods & 10\% & 20\% & 30\% & 40\% & 50\% & 60\% & 70\% & 80\% & 90\% & 95\% & Mean \\
		\hline \hline 
		\multirow{8}{*}{{\rotatebox[origin=c]{90}{20TTI}}} & ADKIF & 40.64 & 37.63 & 35.61 & 33.86 & 32.12 & 30.30 & 28.47 & 26.66 & 24.68 & 23.09 & 31.31 \\
		& NLSF-MLP & 38.95 & 34.74 & 31.95 & 29.79 & 27.96 & 26.24 & 26.62 & 24.25 & 20.85 & 19.51 & 28.09 \\
		& NLSF-CNN & 40.39 & 36.03 & 33.13 & 31.67 & 29.72 & 27.89 & 27.29 & 24.87 & 21.38 & 20.00 & 29.24 \\
		& ARmF & 41.29 & 38.10 & 35.98 & 34.20 & 32.54 & 30.89 & 29.20 & 27.34 & 24.83 & 22.90 & 31.73 \\
		& ACmF & 40.93 & 37.80 & 35.72 & 34.00 & 32.40 & 30.82 & 29.19 & 27.39 & 24.92 & 22.89 & 31.61 \\
		& IAWMF & 41.39 & 38.26 & 36.17 & 34.43 & 32.81 & 31.20 & 29.61 & 27.96 & 25.57 & 23.54 & 32.09 \\
		& NAHAT & 42.22 & 38.94 & 36.72 & 34.90 & 33.27 & 31.69 & 30.04 & 28.17 & 25.70 & 23.85 & 32.55 \\
		& DAMRmF & 41.36 & 38.23 & 36.14 & 34.40 & 32.82 & 31.21 & 29.63 & 27.97 & 25.57 & 23.42 & 32.07 \\
		& SeConvNet & \textBF{44.30} & \textBF{41.28} & \textBF{39.53} & \textBF{36.45} & \textBF{35.85} & \textBF{34.92} & \textBF{32.85} & \textBF{31.14} & \textBF{28.29} & \textBF{25.88} & \textBF{35.05} \\ 
		\hline 
		\multirow{8}{*}{{\rotatebox[origin=c]{90}{BSD68}}} & ADKIF & 36.72 & 33.63 & 31.68 & 30.12 & 28.65 & 27.17 & 25.69 & 24.24 & 22.68 & 21.48 & 28.21 \\
		& NLSF-MLP & 35.20 & 31.05 & 28.43 & 26.50 & 24.94 & 23.54 & 24.02 & 22.05 & 19.16 & 18.14 & 25.30 \\
		& NLSF-CNN & 36.50 & 32.20 & 29.48 & 28.17 & 26.51 & 25.02 & 24.63 & 22.61 & 19.65 & 18.61 & 26.34 \\		
		& ARmF & 36.76 & 33.56 & 31.54 & 29.93 & 28.47 & 27.06 & 25.61 & 24.05 & 22.04 & 20.52 & 27.95 \\
		& ACmF & 36.22 & 33.05 & 31.09 & 29.55 & 28.18 & 26.84 & 25.47 & 23.96 & 22.03 & 20.51 & 27.69 \\
		& IAWMF & 37.26 & 34.19 & 32.21 & 30.61 & 29.17 & 27.82 & 26.43 & 24.93 & 22.98 & 21.50 & 28.71 \\
		& NAHAT & 38.35 & 34.96 & 32.77 & 31.07 & 29.60 & 28.24 & 26.90 & 25.46 & 23.67 & 22.30 & 29.33 \\
		& DAMRmF & 36.05 & 32.87 & 30.87 & 29.33 & 27.98 & 26.73 & 25.50 & 24.17 & 22.40 & 20.84 & 27.67 \\
		& SeConvNet & \textBF{40.47} & \textBF{37.36} & \textBF{35.13} & \textBF{33.12} & \textBF{30.36} & \textBF{30.41} & \textBF{28.67} & \textBF{27.23} & \textBF{24.83} & \textBF{23.21} & \textBF{31.08} \\ 
		\hline 
	\end{tabular}
}
\end{table*}
\begin{table*}[t]
	\caption{Results of PSNR (dB) for different methods at different noise densities on the CBSD68 color dataset.}
	\label{table:psnr:color}
	\centering
	\resizebox{\textwidth}{!}{
	\begin{tabular}{clrrrrrrrrrrr}
		\hline
		Dataset & Methods & 10\% & 20\% & 30\% & 40\% & 50\% & 60\% & 70\% & 80\% & 90\% & 95\% & Mean \\
		\hline \hline
		\multirow{8}{*}{{\rotatebox[origin=c]{90}{CBSD68}}} & ADKIF & 34.53 & 32.24 & 30.63 & 29.26 & 27.95 & 26.60 & 25.25 & 23.89 & 22.40 & 21.21 & 27.40 \\
		& NLSF-MLP & 33.10 & 29.77 & 27.48 & 25.75 & 24.33 & 23.04 & 23.61 & 21.74 & 18.93 & 17.92 & 24.57 \\
		& NLSF-CNN & 34.32 & 30.87 & 28.50 & 27.37 & 25.86 & 24.49 & 24.21 & 22.29 & 19.41 & 18.38 & 25.57 \\
		& ARmF & 35.11 & 32.62 & 30.87 & 29.42 & 28.07 & 26.74 & 25.37 & 23.85 & 21.90 & 20.38 & 27.43 \\
		& ACmF & 34.63 & 32.15 & 30.46 & 29.07 & 27.79 & 26.53 & 25.23 & 23.78 & 21.88 & 20.38 & 27.19 \\
		& IAWMF & 35.19 & 32.95 & 31.30 & 29.89 & 28.61 & 27.35 & 26.06 & 24.62 & 22.73 & 21.26 & 28.00 \\
		& NAHAT & 36.63 & 33.97 & 32.08 & 30.54 & 29.18 & 27.90 & 26.62 & 25.23 & 23.47 & 22.10 & 28.77 \\
		& DAMRmF & 33.92 & 31.49 & 29.85 & 28.56 & 27.40 & 26.29 & 25.18 & 23.93 & 22.21 & 20.62 & 26.94 \\
		& SeConvNet & \textBF{45.10} & \textBF{42.64} & \textBF{39.70} & \textBF{38.19} & \textBF{34.87} & \textBF{33.82} & \textBF{32.20} & \textBF{29.82} & \textBF{26.64} & \textBF{24.38} & \textBF{34.73} \\
		\hline
	\end{tabular}
	}
\end{table*}
\begin{table*}[t]
	\caption{Results of SSIM for different methods at different noise densities on the 20 traditional test images and BSD68 gray-scale datasets.}
	\label{table:ssim:gray}
	\centering
	\resizebox{\textwidth}{!}{
	\begin{tabular}{clrrrrrrrrrrr}
		\hline 
		Dataset & Methods & 10\% & 20\% & 30\% & 40\% & 50\% & 60\% & 70\% & 80\% & 90\% & 95\% & Mean \\
		\hline \hline 
		\multirow{8}{*}{{\rotatebox[origin=c]{90}{20TTI}}} & ADKIF & 0.987 & 0.973 & 0.958 & 0.940 & 0.916 & 0.883 & 0.836 & 0.772 & 0.685 & 0.616 & 0.857 \\
		& NLSF-MLP & 0.946 & 0.899 & 0.860 & 0.827 & 0.798 & 0.765 & 0.782 & 0.702 & 0.579 & 0.521 & 0.768 \\
		& NLSF-CNN & 0.981 & 0.932 & 0.892 & 0.879 & 0.848 & 0.813 & 0.802 & 0.720 & 0.593 & 0.534 & 0.799 \\		
		& ARmF & 0.988 & 0.976 & 0.962 & 0.945 & 0.924 & 0.897 & 0.861 & 0.810 & 0.725 & 0.647 & 0.873 \\
		& ACmF & 0.987 & 0.974 & 0.959 & 0.942 & 0.921 & 0.893 & 0.857 & 0.807 & 0.723 & 0.645 & 0.871 \\
		& IAWMF & 0.988 & 0.976 & 0.962 & 0.945 & 0.924 & 0.897 & 0.862 & 0.812 & 0.743 & 0.672 & 0.878 \\
		& NAHAT & 0.989 & 0.977 & 0.964 & 0.948 & 0.928 & 0.904 & 0.871 & 0.824 & 0.746 & 0.677 & 0.883 \\
		& DAMRmF & 0.988 & 0.976 & 0.961 & 0.945 & 0.924 & 0.897 & 0.862 & 0.813 & 0.744 & 0.671 & 0.878 \\
		& SeConvNet & \textBF{0.995} & \textBF{0.991} & \textBF{0.986} & \textBF{0.975} & \textBF{0.972} & \textBF{0.963} & \textBF{0.948} & \textBF{0.927} & \textBF{0.882} & \textBF{0.833} & \textBF{0.947} \\ 
		\hline
		\multirow{8}{*}{{\rotatebox[origin=c]{90}{BSD68}}} & ADKIF & 0.983 & 0.964 & 0.943 & 0.918 & 0.886 & 0.841 & 0.780 & 0.700 & 0.597 & 0.524 & 0.814 \\
		& NLSF-MLP & 0.942 & 0.890 & 0.846 & 0.808 & 0.771 & 0.729 & 0.729 & 0.637 & 0.504 & 0.443 & 0.730 \\
		& NLSF-CNN & 0.977 & 0.923 & 0.877 & 0.859 & 0.820 & 0.774 & 0.748 & 0.653 & 0.517 & 0.454 & 0.760 \\
		& ARmF & 0.983 & 0.966 & 0.948 & 0.925 & 0.897 & 0.861 & 0.813 & 0.747 & 0.643 & 0.555 & 0.834 \\
		& ACmF & 0.981 & 0.963 & 0.942 & 0.918 & 0.889 & 0.853 & 0.806 & 0.741 & 0.639 & 0.553 & 0.829 \\
		& IAWMF & 0.986 & 0.970 & 0.952 & 0.930 & 0.903 & 0.869 & 0.824 & 0.763 & 0.664 & 0.582 & 0.844 \\
		& NAHAT & 0.988 & 0.973 & 0.955 & 0.932 & 0.905 & 0.870 & 0.824 & 0.761 & 0.665 & 0.588 & 0.846 \\
		& DAMRmF & 0.975 & 0.958 & 0.939 & 0.917 & 0.890 & 0.857 & 0.813 & 0.754 & 0.659 & 0.574 & 0.834 \\
		& SeConvNet & \textBF{0.993} & \textBF{0.987} & \textBF{0.979} & \textBF{0.966} & \textBF{0.946} & \textBF{0.936} & \textBF{0.907} & \textBF{0.875} & \textBF{0.804} & \textBF{0.742} & \textBF{0.913} \\ 
		\hline 
	\end{tabular}
	}
\end{table*}
\begin{table*}[!ht]
	\caption{Results of SSIM for different methods at different noise densities on CBSD68 color dataset.}
	\label{table:ssim:color}
	\centering
	\resizebox{\textwidth}{!}{
	\begin{tabular}{clrrrrrrrrrrr}
		\hline
		Dataset & Methods & 10\% & 20\% & 30\% & 40\% & 50\% & 60\% & 70\% & 80\% & 90\% & 95\% & Mean \\ 
		\hline \hline
		\multirow{8}{*}{{\rotatebox[origin=c]{90}{CBSD68}}} & ADKIF & 0.984 & 0.973 & 0.960 & 0.944 & 0.924 & 0.897 & 0.859 & 0.809 & 0.741 & 0.685 & 0.878 \\
		& NLSF-MLP & 0.944 & 0.898 & 0.861 & 0.831 & 0.804 & 0.777 & 0.803 & 0.736 & 0.626 & 0.579 & 0.786 \\
		& NLSF-CNN & 0.978 & 0.931 & 0.893 & 0.883 & 0.855 & 0.825 & 0.824 & 0.755 & 0.642 & 0.594 & 0.818\\
		& ARmF & 0.985 & 0.975 & 0.964 & 0.950 & 0.933 & 0.910 & 0.879 & 0.836 & 0.762 & 0.692 & 0.889 \\
		& ACmF & 0.984 & 0.973 & 0.960 & 0.946 & 0.928 & 0.905 & 0.875 & 0.832 & 0.760 & 0.692 & 0.885 \\
		& IAWMF & 0.987 & 0.978 & 0.967 & 0.953 & 0.936 & 0.915 & 0.887 & 0.846 & 0.777 & 0.713 & 0.896 \\
		& NAHAT & 0.990 & 0.981 & 0.970 & 0.956 & 0.940 & 0.919 & 0.891 & 0.852 & 0.789 & 0.735 & 0.902 \\
		& DAMRmF & 0.970 & 0.959 & 0.948 & 0.935 & 0.920 & 0.900 & 0.875 & 0.838 & 0.773 & 0.705 & 0.882 \\
		& SeConvNet & \textBF{0.997} & \textBF{0.995} & \textBF{0.992} & \textBF{0.989} & \textBF{0.976} & \textBF{0.971} & \textBF{0.959} & \textBF{0.932} & \textBF{0.869} & \textBF{0.794} & \textBF{0.947} \\ 
		\hline
	\end{tabular}
	}
\end{table*}

Tables \hyperref[table:psnr:gray]{\ref*{table:psnr:gray}} to \hyperref[table:ssim:color]{\ref*{table:ssim:color}} report the SAP denoising performance of the proposed SeConvNet as well as the state-of-the-art competing methods based on the PSNR (dB) and SSIM criteria at varying noise densities ranging from 10\% to 95\%. In addition, the average performance results of 10\% to 95\% noise densities are presented in the last column of the tables. The best result of each noise density is highlighted in bold font. Table \hyperref[table:psnr:gray]{\ref*{table:psnr:gray}} contains the PSNR (dB) results of gray-scale SAP noise reduction on the 20 traditional test images and BSD68 
datasets. The PSNR results of color SAP denoising on CBSD68 
are  listed in table \hyperref[table:psnr:color]{\ref*{table:psnr:color}}. Also included in tables \hyperref[table:ssim:gray]{\ref*{table:ssim:gray}} and \hyperref[table:ssim:color]{\ref*{table:ssim:color}} are the SSIM results of gray-scale and color SAP denoising.

As one can see in the results of the 20 traditional test images, BSD68, and CBSD68 datasets, SeConvNet can attain the best PSNR/SSIM results compared to other counterparts at low, moderate, and high noise densities. SeConvNet surpasses NAHAT, which has the second-best PSNR/SSIM results, by a significant margin. On average, SeConvNet outperforms NAHAT by 2.5dB/\allowbreak 0.064, 1.7dB/\allowbreak 0.067, and 6.0dB/\allowbreak 0.045 in the PSNR/\allowbreak SSIM criterion on the 20 traditional test images, BSD68, and CBSD68 datasets, respectively. Specifically, at very high noise density, i.e., 95\%, while NAHAT can denoise the images in the 20 traditional test images, BSD68, and CBSD68 datasets with a PSNR/\allowbreak SSIM performance of about 23.8dB/\allowbreak 0.883, 22.3dB/\allowbreak 0.846, 22.10dB/\allowbreak 0.735, SeConvNet can drastically boost PSNR/\allowbreak SSIM gain on these datasets to 25.8dB/\allowbreak 0.947, 23.2dB/\allowbreak 0.913, and 24.38dB/\allowbreak 0.794. Aside from SeConvNet and NAHAT, IAWMF can offer the best performance among other remaining methods on these datasets.


\begin{figure*}[t]
	\centering
	\subfloat[]{\label{fig:Peppers:original}
		\includegraphics[width=0.24\textwidth]{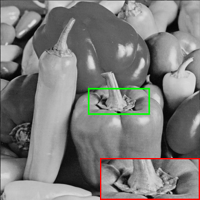}}
	\subfloat[]{\label{fig:Peppers:noisy}
		\includegraphics[width=0.24\textwidth]{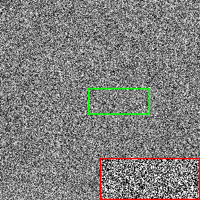}}
	\subfloat[]{\label{fig:Peppers:dba}
		\includegraphics[width=0.24\textwidth]{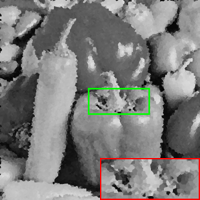}}
	\subfloat[]{\label{fig:Peppers:mdbutmf}
		\includegraphics[width=0.24\textwidth]{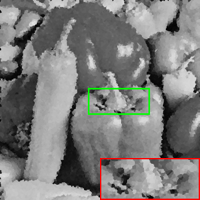}}
	\\[-2ex]
	\subfloat[]{\label{fig:Peppers:awmf}
		\includegraphics[width=0.24\textwidth]{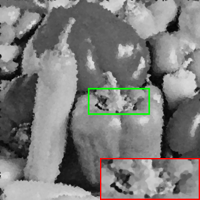}}
	\subfloat[]{\label{fig:Peppers:adkif}
		\includegraphics[width=0.24\textwidth]{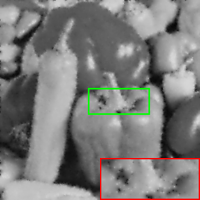}}
	\subfloat[]{\label{fig:Peppers:damf}
		\includegraphics[width=0.24\textwidth]{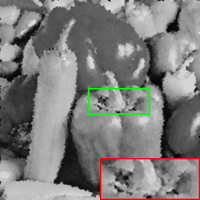}}
	\subfloat[]{\label{fig:Peppers:armf}
		\includegraphics[width=0.24\textwidth]{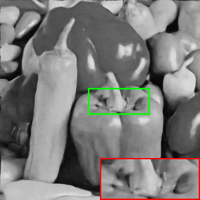}}

	\caption{Restoration results of different methods for the Peppers image with $95\%$ SAP noise. (a)~Original image (b)~Noisy image (c)~ARmF (d)~ACmF (e)~IAWMF (f)~NAHAT (g)~DAMRmF (h)~SeConvNet.}
	\label{fig:Peppers}
\end{figure*}

\begin{figure*}[!t]
	\centering
	\subfloat[]{\label{fig:Parrot:original}
		\includegraphics[width=0.24\textwidth]{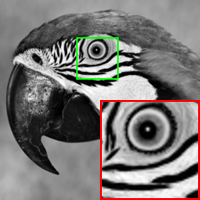}}
	\subfloat[]{\label{fig:Parrot:noisy}
		\includegraphics[width=0.24\textwidth]{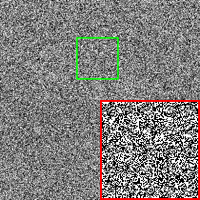}}
	\subfloat[]{\label{fig:Parrot:dba}
		\includegraphics[width=0.24\textwidth]{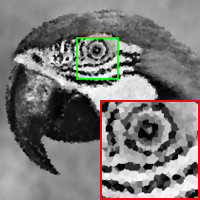}}
	\subfloat[]{\label{fig:Parrot:mdbutmf}
		\includegraphics[width=0.24\textwidth]{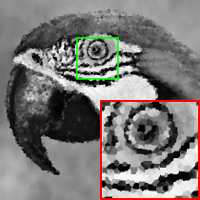}}
	\\[-2ex]
	\subfloat[]{\label{fig:Parrot:awmf}
		\includegraphics[width=0.24\textwidth]{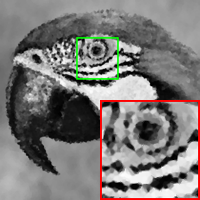}}
	\subfloat[]{\label{fig:Parrot:adkif}
		\includegraphics[width=0.24\textwidth]{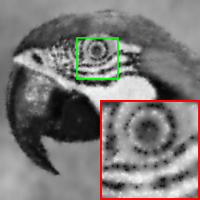}}
	\subfloat[]{\label{fig:Parrot:damf}
		\includegraphics[width=0.24\textwidth]{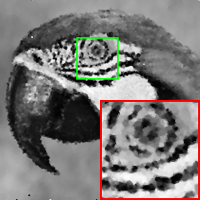}}
	\subfloat[]{\label{fig:Parrot:armf}
		\includegraphics[width=0.24\textwidth]{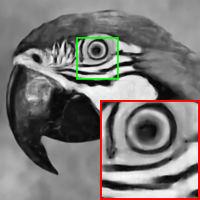}}
	
	\caption{Restoration results of different methods for the Parrot image with $95\%$ SAP noise. (a)~Original image (b)~Noisy image (c)~ARmF (d)~ACmF (e)~IAWMF (f)~NAHAT (g)~DAMRmF (h)~SeConvNet.}
	\label{fig:Parrot}
\end{figure*}

\begin{figure*}[!t]
	\centering
	\subfloat[]{\label{fig:billiard_balls_b:original}
		\includegraphics[width=0.24\textwidth]{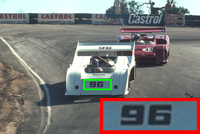}}
	\subfloat[]{\label{fig:billiard_balls_b:noisy}
		\includegraphics[width=0.24\textwidth]{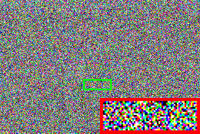}}
	\subfloat[]{\label{fig:billiard_balls_b:dba}
		\includegraphics[width=0.24\textwidth]{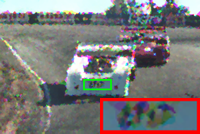}}
	\subfloat[]{\label{fig:billiard_balls_b:mdbutmf}
		\includegraphics[width=0.24\textwidth]{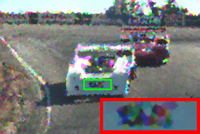}}
	\\[-2ex]
	\subfloat[]{\label{fig:billiard_balls_b:awmf}
		\includegraphics[width=0.24\textwidth]{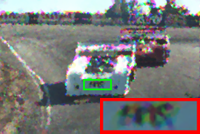}}
	\subfloat[]{\label{fig:billiard_balls_b:adkif}
		\includegraphics[width=0.24\textwidth]{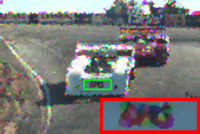}}
	\subfloat[]{\label{fig:billiard_balls_b:damf}
		\includegraphics[width=0.24\textwidth]{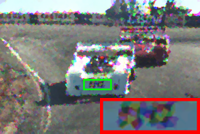}}
	\subfloat[]{\label{fig:billiard_balls_b:armf}
		\includegraphics[width=0.24\textwidth]{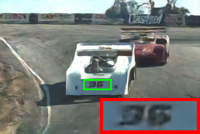}}
	
	\caption{Restoration results of different methods for one image from the CBSD68 dataset with $95\%$ SAP noise. (a)~Original image (b)~Noisy image (c)~ARmF (d)~ACmF (e)~IAWMF (f)~NAHAT (g)~DAMRmF (h)~SeConvNet.}
	\label{fig:21077}
\end{figure*}

\begin{figure*}[!t]
	\centering
	\subfloat[]{\label{fig:253027:original}
		\includegraphics[width=0.24\textwidth]{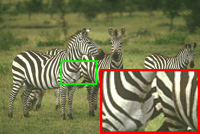}}
	\subfloat[]{\label{fig:253027:noisy}
		\includegraphics[width=0.24\textwidth]{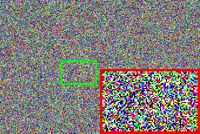}}
	\subfloat[]{\label{fig:253027:dba}
		\includegraphics[width=0.24\textwidth]{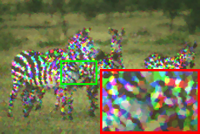}}
	\subfloat[]{\label{fig:253027:mdbutmf}
		\includegraphics[width=0.24\textwidth]{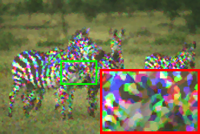}}
	\\[-2ex]
	\subfloat[]{\label{fig:253027:awmf}
		\includegraphics[width=0.24\textwidth]{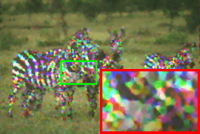}}
	\subfloat[]{\label{fig:253027:adkif}
		\includegraphics[width=0.24\textwidth]{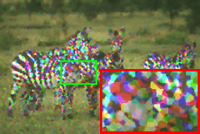}}
	\subfloat[]{\label{fig:253027:damf}
		\includegraphics[width=0.24\textwidth]{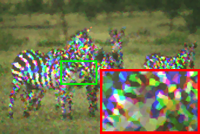}}
	\subfloat[]{\label{fig:253027:armf}
		\includegraphics[width=0.24\textwidth]{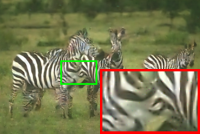}}
	
	\caption{Restoration results of different methods for one image from the CBSD68 dataset with $95\%$ SAP noise. (a)~Original image (b)~Noisy image (c)~ARmF (d)~ACmF (e)~IAWMF (f)~NAHAT (g)~DAMRmF (h)~SeConvNet.}
	\label{fig:253027}
\end{figure*}

\begin{figure*}[!t]
	\centering
	\subfloat[]{\label{fig:test003:original}
		\includegraphics[width=0.24\textwidth,height=3.8cm]{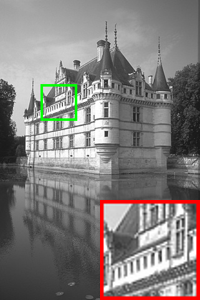}}
	\subfloat[]{\label{fig:test003:noisy}
		\includegraphics[width=0.24\textwidth,height=3.8cm]{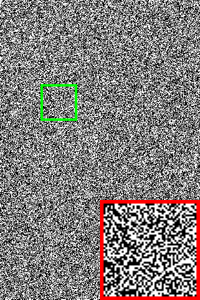}}
	\subfloat[]{\label{fig:test003:dba}
		\includegraphics[width=0.24\textwidth,height=3.8cm]{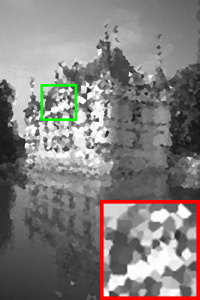}}
	\subfloat[]{\label{fig:test003:mdbutmf}
		\includegraphics[width=0.24\textwidth,height=3.8cm]{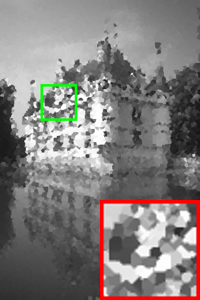}}
	\\[-2ex]
	\subfloat[]{\label{fig:test003:awmf}
		\includegraphics[width=0.24\textwidth,height=3.8cm]{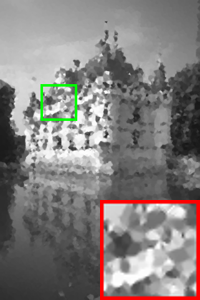}}
	\subfloat[]{\label{fig:test003:adkif}
		\includegraphics[width=0.24\textwidth,height=3.8cm]{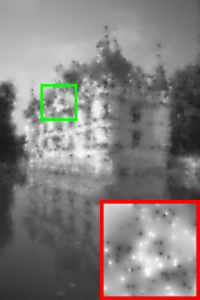}}
	\subfloat[]{\label{fig:test003:damf}
		\includegraphics[width=0.24\textwidth,height=3.8cm]{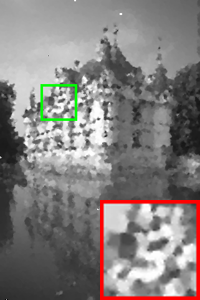}}
	\subfloat[]{\label{fig:test003:armf}
		\includegraphics[width=0.24\textwidth,height=3.8cm]{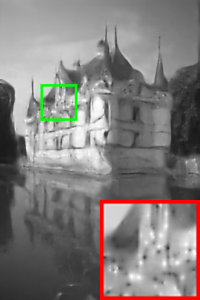}}
	
	\caption{Restoration results of different methods for one image from the BSD68 dataset with $95\%$ SAP noise. (a)~Original image (b)~Noisy image (c)~ARmF (d)~ACmF (e)~IAWMF (f)~NAHAT (g)~DAMRmF (h)~SeConvNet.}
	\label{fig:test003}
\end{figure*}

\begin{figure*}[!t]
	\centering
	\subfloat[]{\label{fig:210088:original}
		\includegraphics[width=0.24\textwidth]{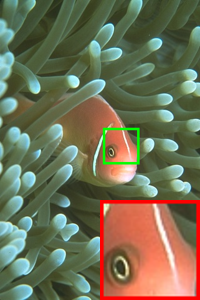}}
	\subfloat[]{\label{fig:210088:noisy}
		\includegraphics[width=0.24\textwidth]{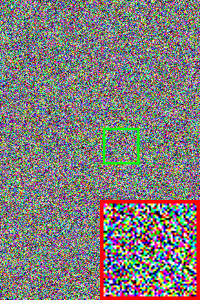}}
	\subfloat[]{\label{fig:210088:dba}
		\includegraphics[width=0.24\textwidth]{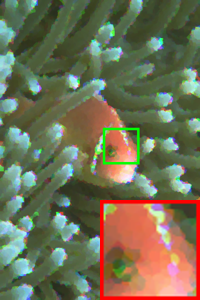}}
	\subfloat[]{\label{fig:210088:mdbutmf}
		\includegraphics[width=0.24\textwidth]{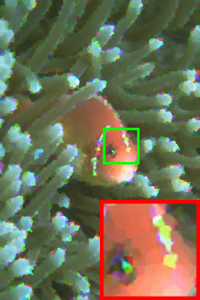}}
	\\[-2ex]
	\subfloat[]{\label{fig:210088:awmf}
		\includegraphics[width=0.24\textwidth]{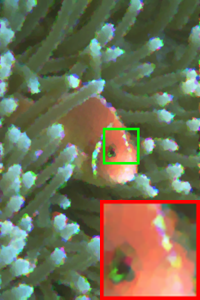}}
	\subfloat[]{\label{fig:210088:adkif}
		\includegraphics[width=0.24\textwidth]{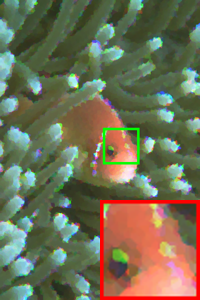}}
	\subfloat[]{\label{fig:210088:damf}
		\includegraphics[width=0.24\textwidth]{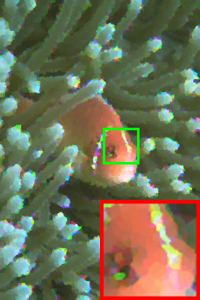}}
	\subfloat[]{\label{fig:210088:armf}
		\includegraphics[width=0.24\textwidth]{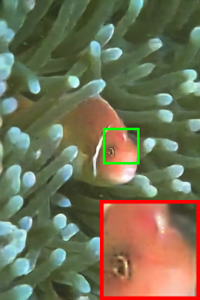}}
	
	\caption{Restoration results of different methods for one image from the CBSD68 dataset with $95\%$ SAP noise. (a)~Original image (b)~Noisy image (c)~ARmF (d)~ACmF (e)~IAWMF (f)~NAHAT (g)~DAMRmF (h)~SeConvNet.}
	\label{fig:210088}
\end{figure*}

To show the SAP denoising efficiency of SeConvNet at a very high noise density, we choose three gray and three color images and illustrate the visual results of various methods on the corrupted version of them by 95\% SAP noise in Figures \hyperref[fig:Peppers]{\ref*{fig:Peppers}} to \hyperref[fig:210088]{\ref*{fig:210088}}. Based on the results shown, it is evident that ARmF, ACmF, IAWMF, and DAMRmF fail to recover the edges and details and produce artifacts in the edges. The NAHAT method blurs images and generates overly smooth textures and edges. However, SeConvNet not only preserves fine details and sharp edges but also {provides impressive visual results in smooth regions}. For instance, it is obviously seen that while other methods fail to recover the eye and the surrounding area of the parrot's eye in figure \hyperref[fig:Parrot]{\ref*{fig:Parrot}}, SeConvNet can effectively preserve fine details and sharp edges, as well as textures. Also, in figure \hyperref[fig:21077]{\ref*{fig:21077}} the number 96 on the white racing car becomes entirely indistinct by other SAP denoising methods, whereas it is almost clearly visible by SeConvNet. Moreover, in the case of color SAP denoising, ARmF, ACmF, IAWMF, NAHAT, and DAMRmF tend to introduce false color artifacts in addition to losing edges and details. These false-color artifacts are clearly apparent in figure \hyperref[fig:253027]{\ref*{fig:253027}}, whilst SeConvNet can recover high-quality visual results without any false-color artifacts. 

For better visual comparison of the obtained results, figures \hyperref[fig:plot:20TTI]{\ref*{fig:plot:20TTI}}, \hyperref[fig:plot:BSD68]{\ref*{fig:plot:BSD68}}, and \hyperref[fig:plot:CBSD68]{\ref*{fig:plot:CBSD68}} demonstrate the line graph of the average PSNR (dB) and SSIM versus noise densities for the 20 traditional test images, BSD68, and CBSD68 dataset, respectively. As it can be obviously seen, the proposed SeConvNet model can consistently surpass other state-of-the-art methods by a considerable margin.

\section{Conclusion}
\label{sec:Conclusion}
A new CNN model, namely SeConvNet, is proposed in this paper to denoise both gray-scale and color images corrupted  by SAP noise, especially at high and very high noise densities. To meet this purpose, we introduced a new selective convolutional (SeConv) block for the beginning part of the network, which computes an initial estimation of noisy pixels by considering only non-noisy pixels. In the following layers of the network, conventional convolutional layers, as well as batch normalization (BN) and rectified linear units (ReLU), are employed. The denoising performance of SeConvNet is compared with seven state-of-the-art methods in terms of the PSNR and SSIM criteria on different gray-scale and color datasets, including 20 traditional test images and gray and color versions of 68 images of the Berkeley segmentation dataset (BSD68 and CBSD68) 
. The results indicate that SeConvNet not only surpasses all counterparts by a significant margin (particularly at high noise densities) in quantitative denoising performance but also produces favorable restored images by preserving fine details and sharp edges along with providing impressive visual results in the smooth region.

\begin{figure*}[!t]
	\centering
	\subfloat[]{\label{fig:plot:20TTI:psnr}
		\includegraphics[width=0.48\textwidth, height=0.185\textheight]{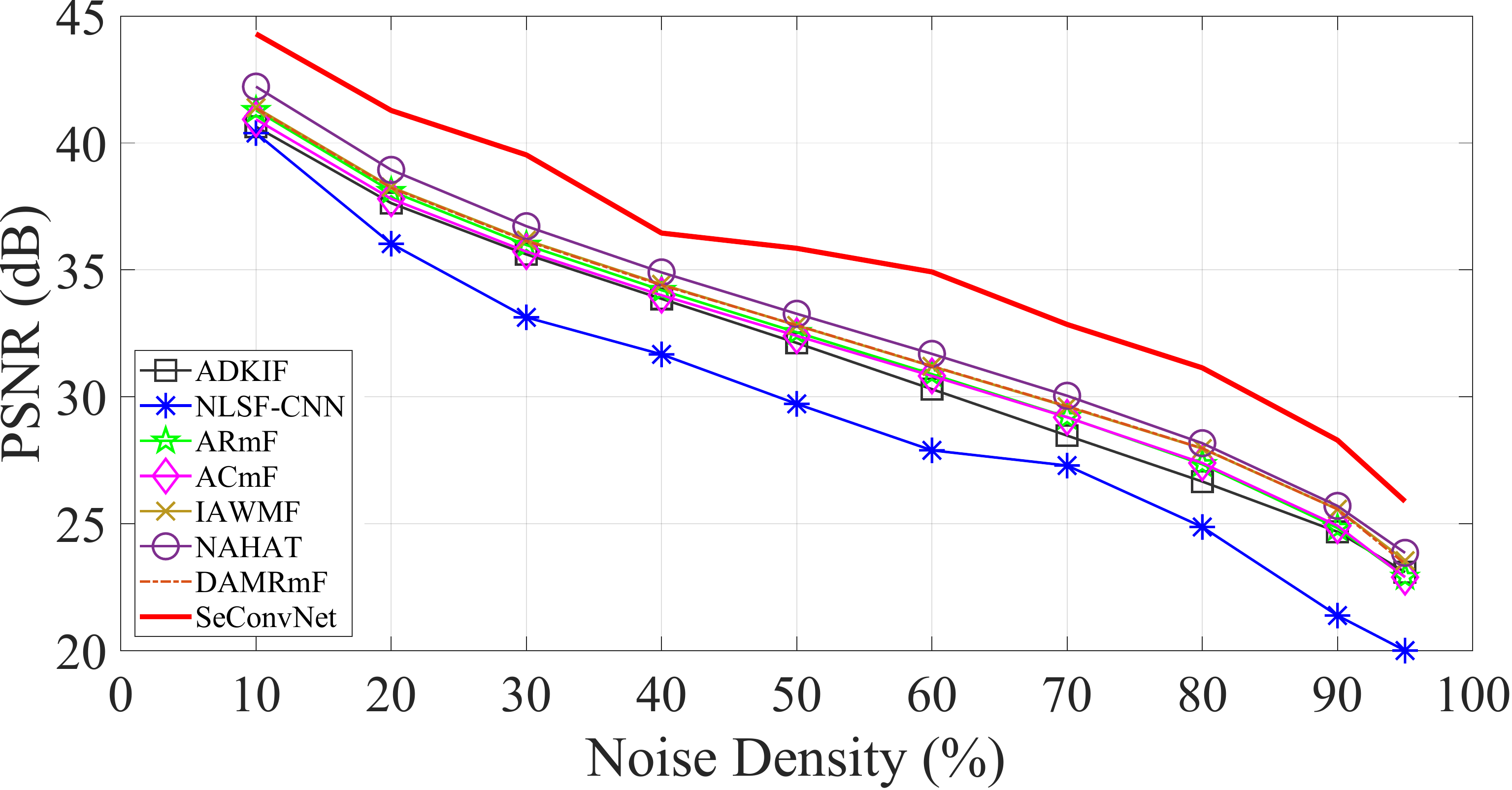}}
	~
	\subfloat[]{\label{fig:plot:20TTI:ief}
		\includegraphics[width=0.48\textwidth, height=0.185\textheight]{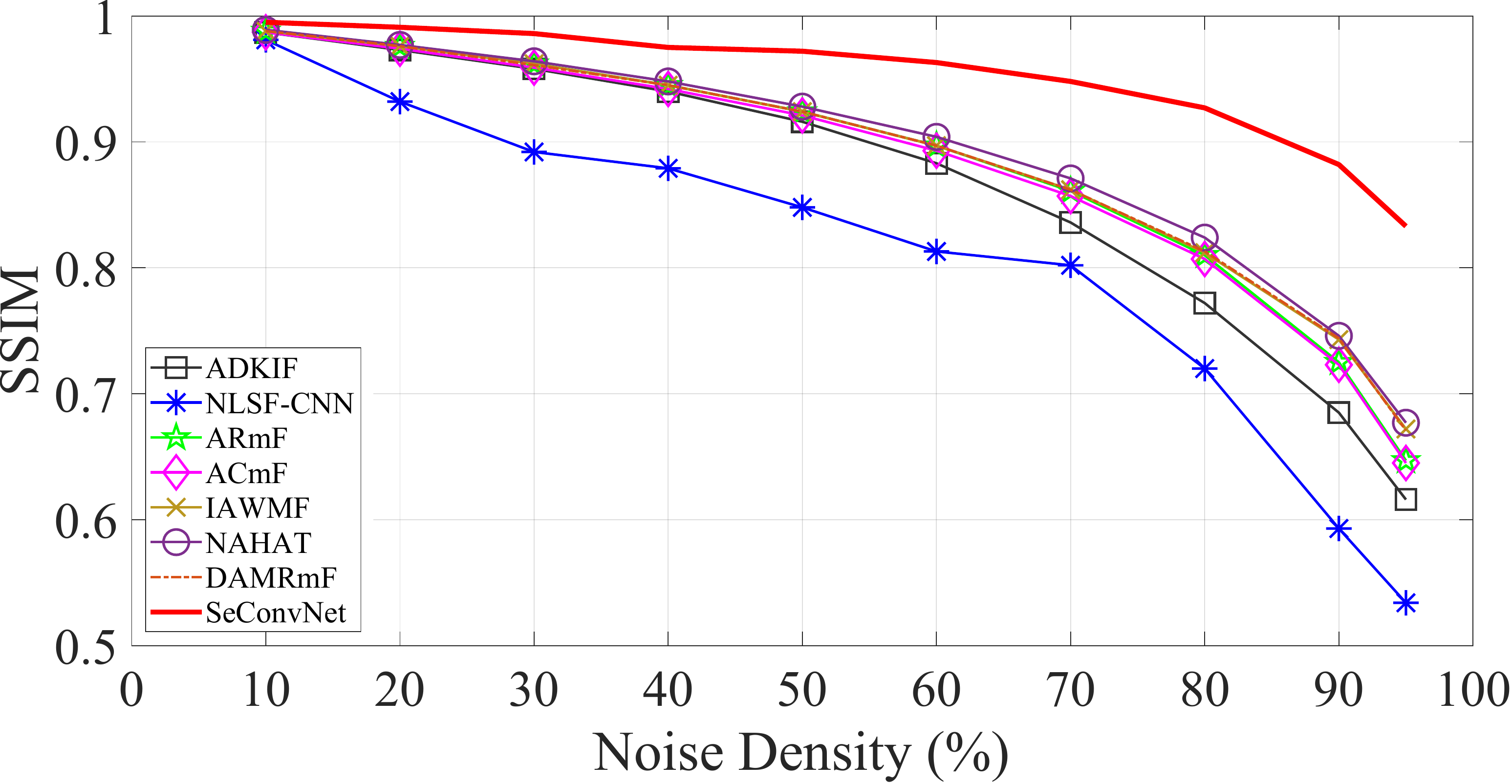}}	
	
	\caption{Comparison graphs of the averages of different metrics on the 20 traditional test images dataset versus different noise densities. (a)~PSNR~(dB) (b)~SSIM.}
	\label{fig:plot:20TTI}
\end{figure*}
\begin{figure*}[!t]
	\centering
	\subfloat[]{\label{fig:plot:BSD68:psnr}
		\includegraphics[width=0.48\textwidth, height=0.185\textheight]{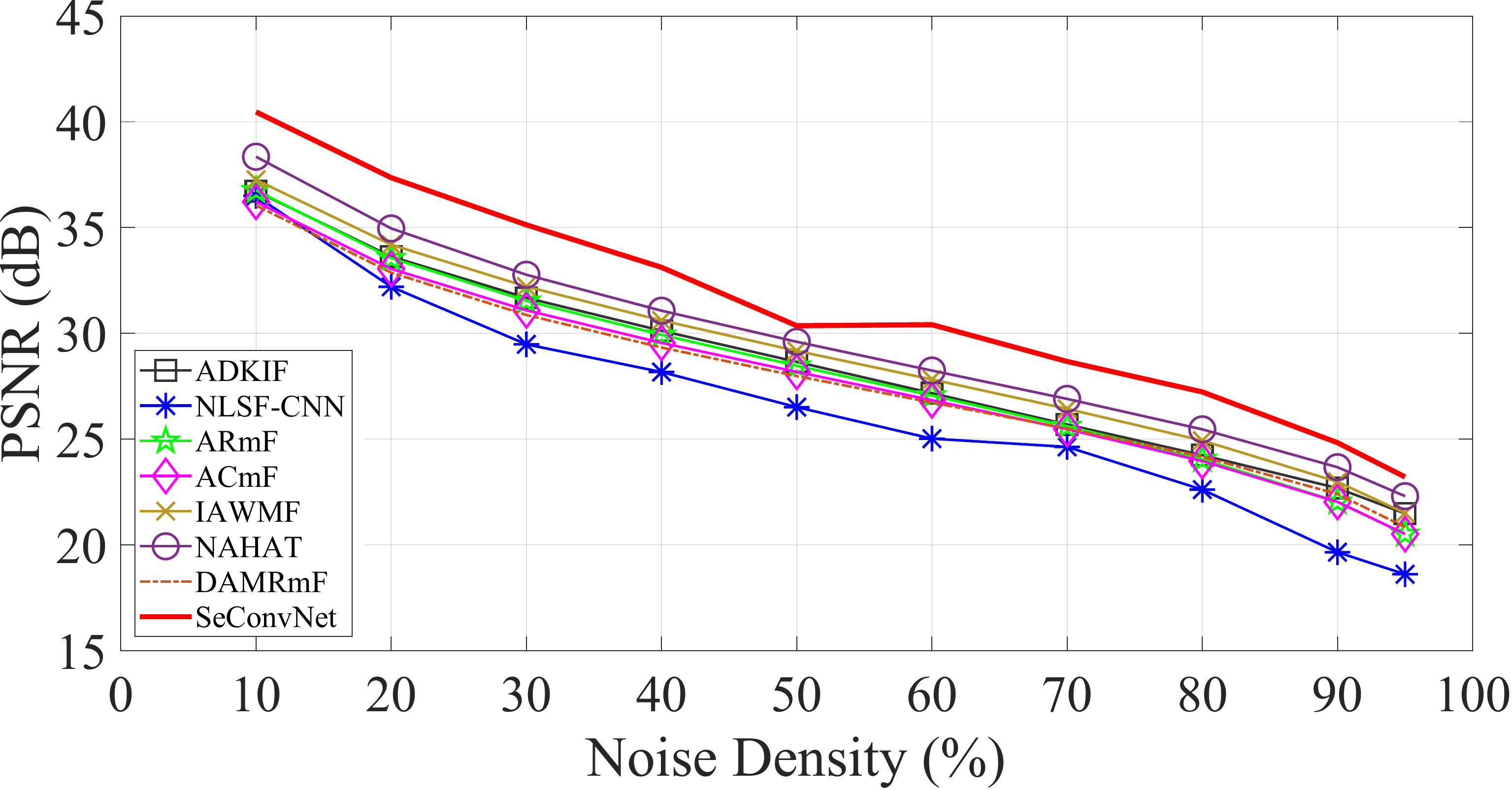}}
	~
	\subfloat[]{\label{fig:plot:BSD68:ief}
		\includegraphics[width=0.48\textwidth, height=0.185\textheight]{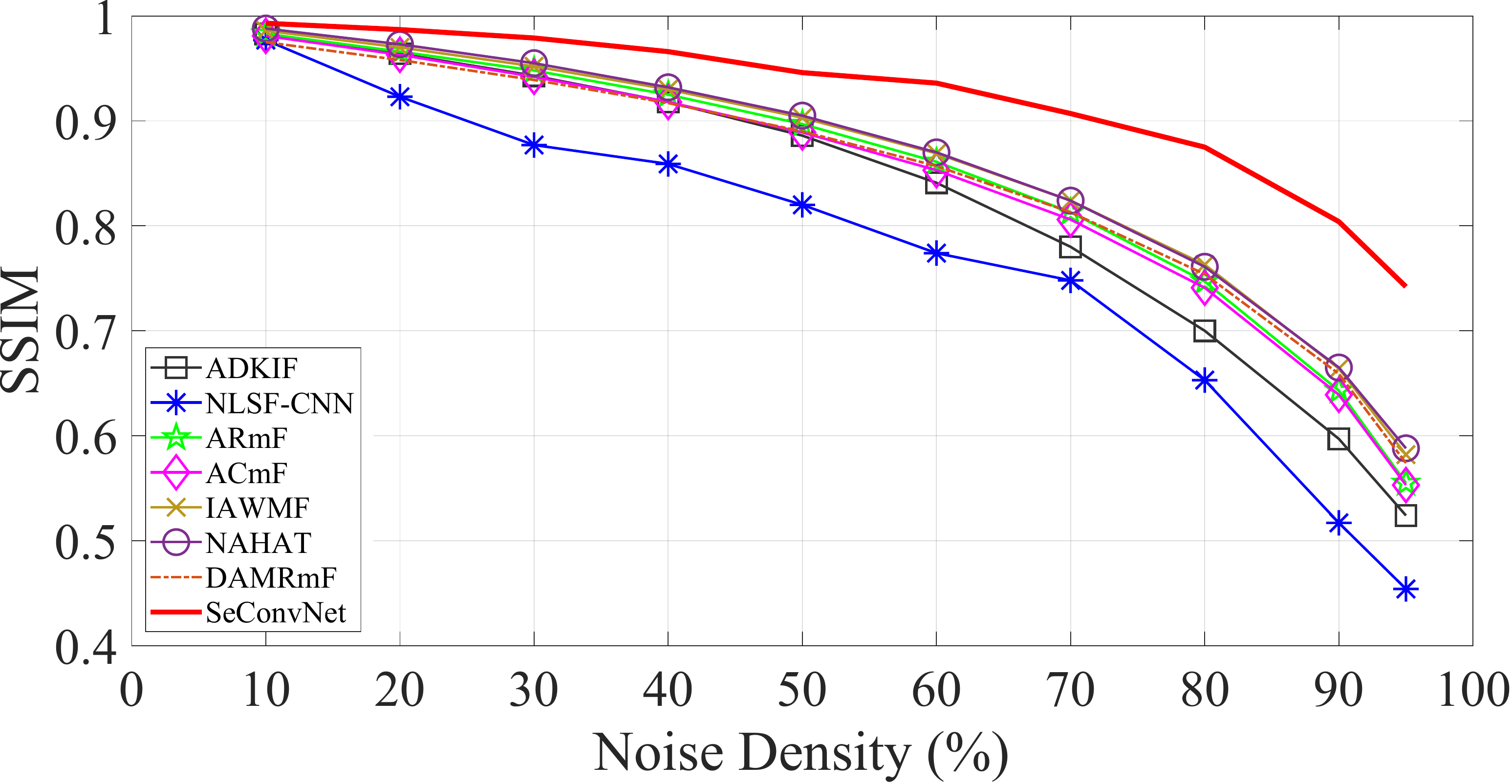}}	
	
	\caption{Comparison graphs of the averages of different metrics on the BSD68 dataset versus different noise densities. (a)~PSNR~(dB) (b)~SSIM.}
	\label{fig:plot:BSD68}
\end{figure*}
\begin{figure*}[!t]
	\centering
	\subfloat[]{\label{fig:plot:CBSD68:psnr}
		\includegraphics[width=0.48\textwidth, height=0.185\textheight]{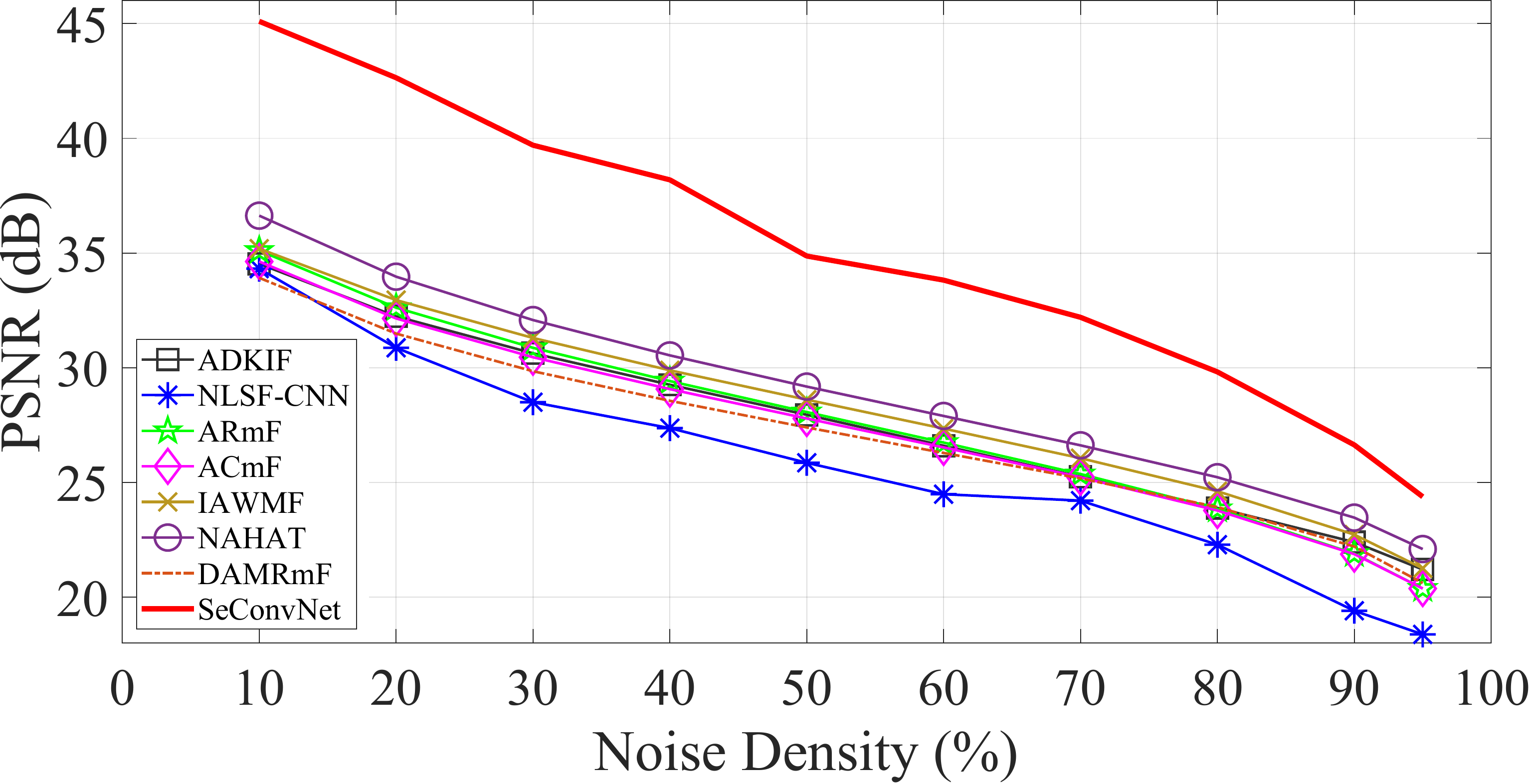}}
	~
	\subfloat[]{\label{fig:plot:CBSD68:ief}
		\includegraphics[width=0.48\textwidth, height=0.185\textheight]{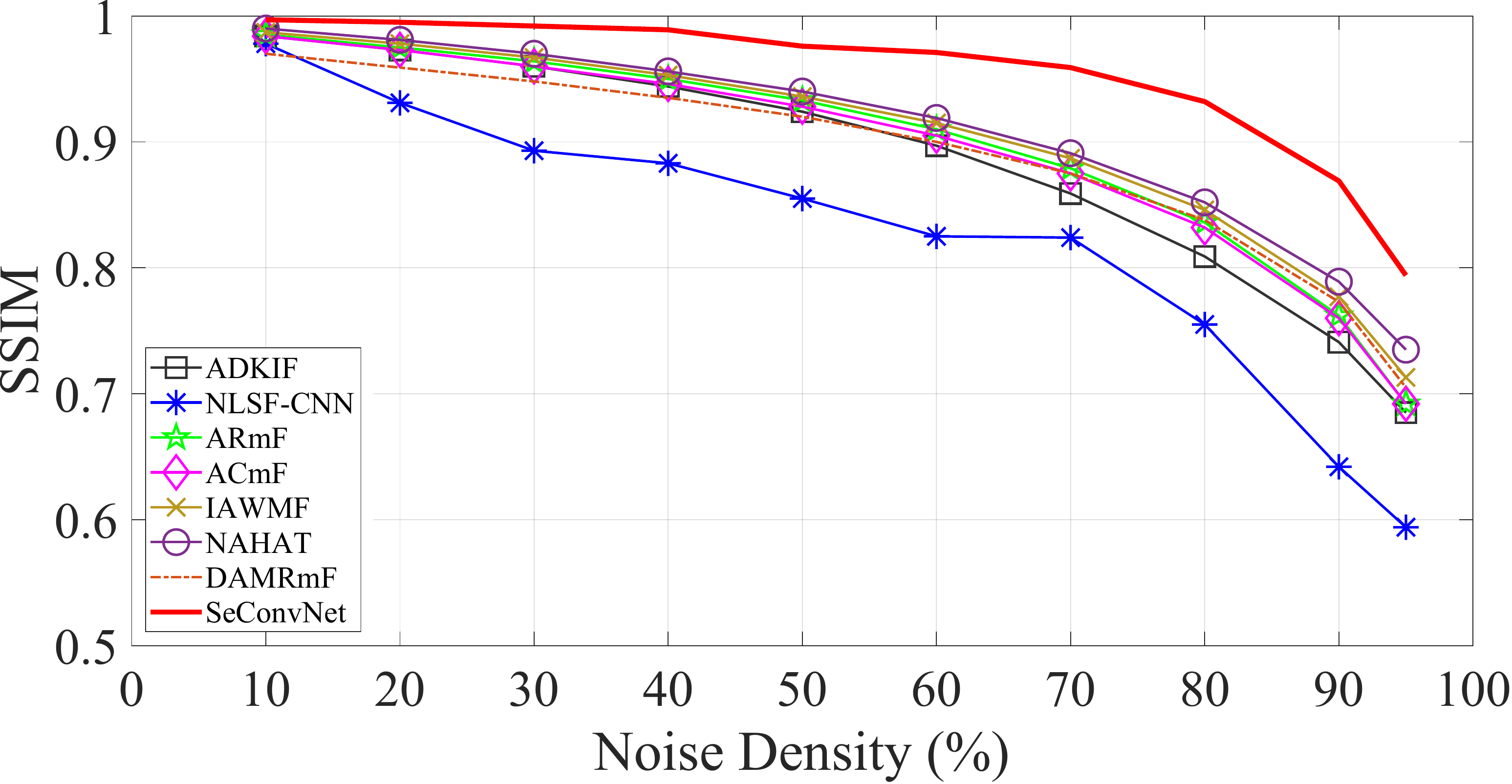}}	
	
	\caption{Comparison graphs of the averages of different metrics on the CBSD68 dataset versus different noise densities. (a)~PSNR~(dB) (b)~SSIM.}
	\label{fig:plot:CBSD68}
\end{figure*}

\bibliographystyle{unsrtnat}
\bibliography{refs}  






\end{document}